    \pdfoutput=1
    \documentclass[11pt]{article}

    \usepackage{ACL2023}
    \usepackage{custom}
    \usepackage{anyfontsize}

\newcommand{\mymacro}[1]{{#1}}

\newcommand{\ethz}{\emoji[emojis]{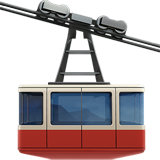}}
\newcommand{\cop}{\emoji[emojis]{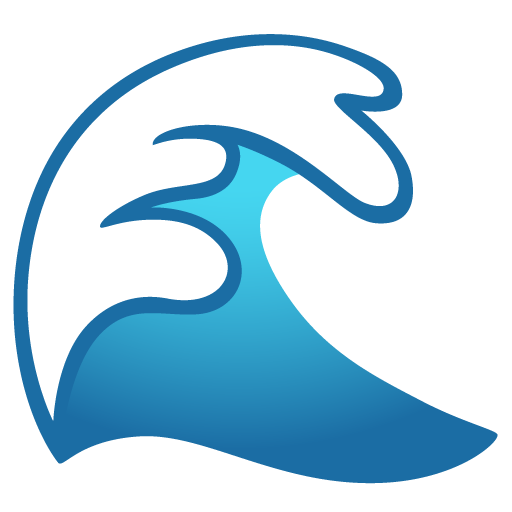}}
\newcommand{\upenn}{\emoji[emojis]{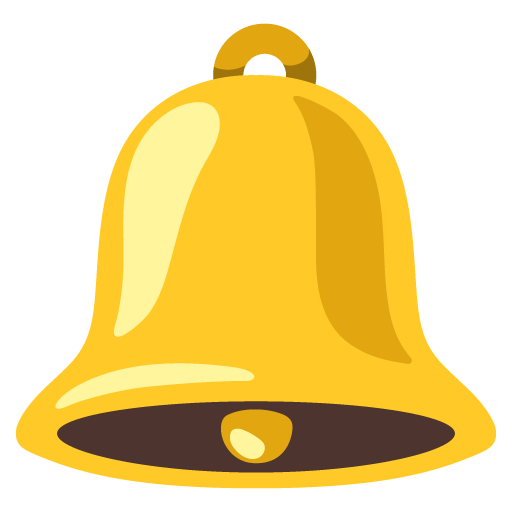}}

\newcommand{\defn}[1]{\textbf{#1}}

\newcommand{\paroutline}[3][false]{%
    \ifnum\pdfstrcmp{#1}{true}=0
        #3% Print only the second argument
    \else
        [\textit{\textcolor{DiverseMagenta}{#2}}] \textcolor{AccentBlue}{#3}% Print both arguments
    \fi
}

\newcommand{\meanStd}[2]{{\mymacro{\shortstack{$#1$  \\  \textcolor{ETHGray}{$\scriptstyle \left(\pm #2\right)$}}}}}
\newcommand{\meanStdSameRow}[2]{{\mymacro{$#1$ \textcolor{ETHGray}{$\scriptstyle \left(\pm #2\right)$}}}}

\newcommand{\pdens}{{\mymacro{ p}}}

\newcommand{\qdens}{{\mymacro{ q}}}

\newcommand{\R}{{\mymacro{ \mathbb{R}}}}

\newcommand{\vfunc}{{\mymacro{ \boldsymbol{f}}}}

\newcommand{\alphabet}{{\mymacro{ \Sigma}}}

\newcommand{\eosalphabet}{{\mymacro{ \overline{\alphabet}}}}
\newcommand{\bosalphabet}{{\mymacro{ \underline{\alphabet}}}}

\newcommand{\kleene}[1]{{\mymacro{#1^*}}}

\newcommand{\str}{{\mymacro{\boldsymbol{y}}}}

\newcommand{\strlt}{{\mymacro{ \str_{<\tstep}}}}

\newcommand{\strlen}{{\mymacro{T}}}

\newcommand{\sym}{{\mymacro{y}}}
\newcommand{\eossym}{{\mymacro{\overline{\sym}}}}

\newcommand{\defeq}{\mathrel{\stackrel{\textnormal{\tiny def}}{=}}}

\newcommand{\set}[1]{{\mymacro{\left\{ #1 \right\}}}}

\newcommand{\bigo}{{\mymacro{ \mathcal{O}}}}

\newcommand{\idxi}{{\mymacro{ i}}}
\newcommand{\idxj}{{\mymacro{ j}}}

\newcommand{\nsymbols}{{\mymacro{ |\alphabet|}}}
\newcommand{\eosnsymbols}{{\mymacro{ |\eosalphabet|}}}
\newcommand{\bosnsymbols}{{\mymacro{ |\bosalphabet|}}}

\newcommand{\tstep}{{\mymacro{ t}}}

\newcommand{\pLM}{\mymacro{\pdens}}
\newcommand{\pLMn}{\mymacro{\pdens_\ngr}}

\newcommand{\qLM}{\mymacro{\qdens}}
\newcommand{\pLNSM}{\mymacro{\pdens}}

\newcommand{\bos}{{\mymacro{\textsc{bos}}}}
\newcommand{\eos}{{\mymacro{\textsc{eos}}}}
\newcommand{\ngr}{{\mymacro{ \textit{n}}}}
\newcommand{\nhat}{{\mymacro{\widehat{\ngr}}}}
\newcommand{\ngram}{{\mymacro{ \textit{n}-gram}}\xspace}

\newcommand{\rank}{{\mymacro{R}}}

\newcommand{\onehot}[1]{{\mymacro{ \llbracket#1\rrbracket}}}

\newcommand{\inEmbedding}{{\mymacro{ \vr}}}
\newcommand{\inEmbeddingFun}[2][]{{\mymacro{ \inEmbedding\!\left(#2\right)}}}

\newcommand{\symt}{{\mymacro{ \sym_{\tstep}}}}

\newcommand{\one}{{\mymacro{\mathbf{1}}}}

\newcommand{\outMtx}{{\mymacro{ \mE}}}

\newcommand{\hiddDim}{{\mymacro{ D}}}

\newcommand{\repFunc}{\mymacro{r}}
\newcommand{\repFuncFun}[1]{\mymacro{\repFunc}\left(#1\right)}

\newcommand{\height}[1]{\textsf{height}(#1)}

\newcommand{\enc}{{\mymacro{\boldsymbol{h}}}}
\newcommand{\encfunc}[1]{\enc (#1)}

\newcommand{\mlp}{{\mymacro{\vfunc}}}

\newcommand{\negterm}[1]{{\mymacro{ {\raise.17ex\hbox{$\scriptstyle\sim$}} #1}}}

\newcommand{\ignore}[1]{}
\newcommand{\expandLater}[1]{}

\newcommand{\tfheadnum}{\mymacro{H}}

\newcommand{\transformernetwork}{\mymacro{\mathcal{T}}}

\newcommand{\tfpLM}{\mymacro{\pLM_\transformernetwork}}

\newcommand{\tfnumlayer}{\mymacro{L}}

\def\1{\mathbf{1}}
\newcommand{\dataset}{{\mymacro{\mathcal{D}}}}
\newcommand{\datasetTrain}{{\mymacro{\dataset_{\text{Train}}}}}

\newcommand{\datasetTest}{{\mymacro{\dataset_{\text{Test}}}}}
\newcommand{\datasetSize}{{\mymacro{N}}}
\newcommand{\datasetSizeTrain}{{\mymacro{\datasetSize_{\text{Train}}}}}

\newcommand{\datasetSizeTest}{{\mymacro{\datasetSize_{\text{Test}}}}}

\def\vr{{{\mymacro{ \mathbf{r}}}}}

\def\vx{{{\mymacro{ \mathbf{x}}}}}

\def\evx{{{\mymacro{ x}}}}

\def\mE{{{\mymacro{ \mathbf{E}}}}}

\newcommand{\E}{{\mymacro{ \mathbb{E}}}}

\newcommand{\N}{{\mymacro{ \mathbb{N}}}}

\newcommand{\KL}{{\mymacro{ D_{\mathrm{KL}}}}}
\newcommand{\KLFun}[2]{\KL\left(#1 \mid \mid #2\right)}

\newcommand{\entropy}{{\mymacro{ \mathrm{H}}}}

\newcommand{\softmax}{{\mymacro{ \mathrm{softmax}}}}
\newcommand{\sparsemax}{{\mymacro{ \mathrm{sparsemax}}}}

\newcommand{\softmaxfunc}[2]{{\mymacro{ \mathrm{softmax}\!\left(#1\right)_{#2}}}} % vector, index}
\newcommand{\bigO}[1]{{\mymacro{ \mathcal{O}\left(#1\right)}}}

\newcommand{\smoothed}[1]{\mymacro{\tilde{#1}}}

\newcommand{\qngram}{\mymacro{q_{\mathrm{MLE}}^n}}

\newcommand{\Count}{\mymacro{\#}}

\newcommand{\CountFun}[1]{\mymacro{\Count\mleft(#1\mright)}}
\newcommand{\typecount}{\mymacro{\Count^n_{\textsc{t}}}}

\newcommand{\CountAS}{\mymacro{\Count_{\mathrm{AS}}}}

\newcommand{\pAS}{\mymacro{\smoothed{q}^n_{\mathrm{AS}}}}
\newcommand{\pWB}{\mymacro{\smoothed{q}^n_{\mathrm{WB}}}}
\newcommand{\pWBminus}{\mymacro{\smoothed{q}^{n-1}_{\mathrm{WB}}}}

\newcommand{\pAD}{\mymacro{\smoothed{q}^n_{\mathrm{AD}}}}
\newcommand{\pADminus}{\mymacro{\smoothed{q}^{n-1}_{\mathrm{AD}}}}

\newcommand{\nstr}{\mymacro{\str^{n}}}

\newcommand{\nmstr}{\mymacro{\str^{n-1}}}

    \title{Can Transformers Learn \ngram Language Models?}
    
    \author{
     Anej Svete$^{\ethz}$ ~\;~
     Nadav Borenstein$^{\cop}$\\
     \textbf{Mike Zhou}$^{\upenn}$ ~\;~
        \textbf{Isabelle Augenstein}$^{\cop}$~\;~
      \textbf{Ryan Cotterell}$^{\ethz}$
    \\
     $^{\ethz}$ETH Zürich~\;~  $^{\cop}$University of Copenhagen~\;~  
     $^{\upenn}$University of Pennsylvania\\
     \{\texttt{\href{mailto:nb@di.ku.dk}{nb}}, \texttt{\href{mailto:augenstein@di.ku.dk} {augenstein}}\}\texttt{@di.ku.dk} \quad 
     \texttt{\href{mailto:mikezhou@seas.upenn.edu}{mikezhou}}\texttt{@seas.upenn.edu} \\
     \{\texttt{\href{mailto:asvete@inf.ethz.ch}{asvete}}, \texttt{\href{mailto:ryan.cotterell@inf.ethz.ch}{ryan.cotterell}}\}\texttt{@inf.ethz.ch}
    }

\begin{document}
    \maketitle
    \begin{abstract}
        Much theoretical work has described the ability of transformers to represent formal languages.
        However, linking theoretical results to empirical performance is not straightforward due to the complex interplay between the architecture, the learning algorithm, and training data.
    To test whether theoretical lower bounds imply \emph{learnability} of formal languages, we turn to recent work relating transformers to \ngram language models (LMs).
        We study transformers' ability to learn random \ngram LMs of two kinds: ones with arbitrary next-symbol probabilities and ones where those are defined with shared parameters.
        We find that classic estimation techniques for \ngram LMs such as add-$\lambda$ smoothing outperform transformers on the former, while transformers perform better on the latter, outperforming methods specifically designed to learn \ngram LMs.
        
        \vspace{0.5em}
        % {\includegraphics[width=1.36em,height=1.25em]{figures/github.png}\hspace{2pt}\parbox{\dimexpr\linewidth-2\fboxsep-2\fboxrule}{\href{https://github.com/rycolab/learning-ngrams}{\texttt{github.com/rycolab/learning-ngrams}}}}
        {\includegraphics[width=1.36em,height=1.25em]{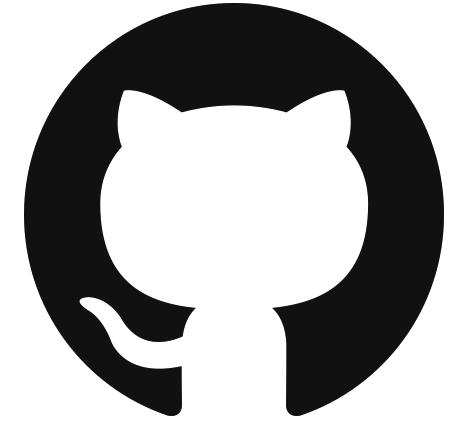}\hspace{2pt}\raisebox{0.5\height}{\parbox{\dimexpr\linewidth-2\fboxsep-2\fboxrule}{\href{https://github.com/rycolab/learning-ngrams}{\texttt{github.com/rycolab/learning-ngrams}}}}}
    \end{abstract}

    \section{Introduction}

    A large body of work has investigated the ability of transformers \citep{Vaswani2017} to represent formal languages \citep{strobl2023transformers}.
    Such results tell us what languages transformers can \emph{represent}, but not how well they can \emph{learn} them from data.
    Existing work has tested the learning abilities of transformers as \emph{classifiers} mapping strings to language membership decisions \citep[e.g.,][]{bhattamishra-etal-2020-on-ability,deletang2023neural}.
    However, language models (LMs) are not classifiers of strings but rather \emph{distributions} over them.
    In this light, a recent line of work has argued for a more direct evaluation of LMs as distributions \citep{svete-cotterell-2023-recurrent,nowak-etal-2023-representational,svete2024theoretical,svete-etal-2024-lower,borenstein2024languages,nowak-etal-2024-computational}.\looseness=-1

    In the pursuit to understand transformers as probability distributions, \defn{\ngram LMs} have recently emerged as a useful playground for analyzing transformers' interpretability \citep{liu2024infinigramscalingunboundedngram,voita-etal-2024-neurons}, learning behavior \citep{edelman2024evolutionstatisticalinductionheads,DBLP:conf/iclr/ChenSCLS24}, in-context learning \citep{olsson2022context,DBLP:conf/iclr/AroraETJP0RR24,akyürek2024incontext,nguyen2024understandingtransformersngramstatistics}, and representational capacity \citep{svete2024transformers}.
    \citet{svete2024transformers}, in particular, show that transformers can encode any \ngram LM, lower-bounding their representational capacity and suggesting that encoding \ngram statistics is indeed a feasible mechanism for transformers to process language. 
    However, being able to \emph{represent} \ngram statistics is insufficient to rely on them in practice.
    For that, \emph{learning} them is necessary.
    This motivates the question: Given that transformers can represent \ngram LMs, how good are they at learning them?
    Answering this brings us closer to a practical understanding of transformers, showing whether \ngram statistics are practically relevant for their behavior and illuminating what theoretical results on neural architectures imply about their ability to learn formal languages.
    
    We study how well transformers can learn \ngram LMs.
    On the one hand, we find that they generalize well on \ngram LMs whose next-symbol probabilities depend on a linear combination of features of the \ngram symbols (representation-based \ngram LMs), outperforming baselines such as count-based techniques and models with hand-crafted features. 
    On the other, we find that transformers perform worse than count-based techniques on \ngram LMs with arbitrary next-symbol probabilities.
    This illuminates transformers' inductive biases: It shows that they struggle to approximate even simple probability distributions whose next-symbol probabilities are arbitrary while being good at learning representation-based LMs.

    \section{Preliminaries}
    We begin by introducing some preliminaries.
    An \defn{alphabet} $\alphabet$ is a finite, non-empty set of \defn{symbols}.
    The \defn{Kleene closure} $\kleene{\alphabet}$ of the alphabet $\alphabet$ is the set of all strings of the symbols in $\alphabet$.
    The \defn{length} of the string $\str = \sym_1\ldots\sym_\strlen \in \kleene{\alphabet}$, denoted by $|\str|=\strlen$, is the number of symbols it contains.
    We will use $\str_{\idxi}^{\idxj} \defeq \sym_{\idxi} \cdots \sym_{\idxj}$ to denote substrings between positions $\idxi$ and $\idxj$ inclusive.

    A \defn{language model} $\pLM$ is a probability distribution over $\kleene{\alphabet}$.
    Two LMs $\pLM$ and $\qLM$ are \defn{weakly equivalent} if $\pLM\left(\str\right) = \qLM\left(\str\right)$ for all $\str \in \kleene{\alphabet}$.
    Many LMs define $\pLM\left(\str\right)$ autoregressively:
    \begin{equation} \label{eq:lnlm}
        \pLM\left(\str\right) \defeq \pLM\left(\eos\mid\str\right) \prod_{\tstep = 1}^{|\str|} \pLM\left(\symt \mid \strlt\right),
    \end{equation}
    where $\eos \notin \alphabet$ is a special \underline{e}nd-\underline{o}f-\underline{s}tring symbol.
    We denote $\eosalphabet \defeq \alphabet \cup \set{\eos}$.

    When defined autoregressively, next-symbol distributions $\pLM\left(\symt \mid \strlt\right)$ in an \ngram{} LM satisfy the \ngram{} assumption, stated formally below.
    \begin{assumption} \label{def:ngram}
        The \defn{\ngram{} assumption} states that the conditional probability of the symbol $\sym_\tstep$ given $\strlt$ only depends on $\ngr-1$ previous symbols:
        % $\str^{\tstep - 1}_{\tstep - \ngr + 1} \defeq \sym_{\tstep-1},\ldots,\sym_{\tstep-\ngr+1}$:\looseness=-1
        \begin{equation} \label{eq:n-gram-assumption}
            \pLMn\left(\sym_\tstep\mid \str_{<\tstep}\right) = \pLMn\left(\sym_\tstep \mid \str^{\tstep - 1}_{\tstep - \ngr + 1}\right).
        \end{equation}
        \noindent We will refer to $\ngr$ as the \defn{order} of the \ngram LM and $\str^{\tstep - 1}_{\tstep - \ngr + 1}$ as the \defn{history} of $\sym_\tstep$.
    \end{assumption}
    We will denote a length-$(\ngr - 1)$ history as $\nstr$ whenever its position $\tstep$ in the string is unimportant.
    Additionally, we assume that the histories are padded with the \underline{b}eginning-\underline{o}f-\underline{s}tring symbol $\bos \notin \alphabet$ and denote $\bosalphabet \defeq \alphabet \cup \set{\bos}$ for convenience.\looseness=-1

    \paragraph{Representation-based \ngram LMs.}
    The next-symbol probabilities $\pLMn\left(\sym\mid \nstr\right)$ of \ngram LMs can be arbitrary.
    This requires storing $\bigo(\nsymbols^{\ngr - 1})$ parameters and does not capture the intuitive notion that similar \ngram{}s define similar next-symbol probabilities.
    This can be addressed by \defn{representation-based} \ngram LMs, which define $\pLMn\left(\sym\mid \nstr\right)$ as a function of a \defn{representation} of $\nstr$.
    \begin{definition}
        An \ngram LM $\pLMn$ is \defn{representation-based} if there exist a \defn{representation function} $\enc\colon \bosalphabet^{\ngr - 1} \to \R^{\hiddDim}$ and an \defn{output matrix} $\outMtx \in \eosnsymbols \times \hiddDim$ for some $\hiddDim \in \N$ such that
        \begin{equation}
            \pLMn\left(\eossym\mid \nstr\right) = \softmax\left(\outMtx \; \enc\left(\nstr\right)\right)_\eossym
        \end{equation}
        for all $\eossym \in \eosalphabet$ and $\nstr \in \bosalphabet^{\ngr - 1}$ where
        \begin{equation}
            \softmaxfunc{\vx}{\eossym} \defeq \frac{\exp(\evx_{\eossym})}{\sum_{\eossym' \in \eosalphabet} \exp(\evx_{\eossym'})}.
        \end{equation}
    \end{definition}

    A transformer LM computes the $\pLM\left(\symt \mid \strlt\right)$ using the self-attention mechanism, which builds the representation by attending to different symbols in the preceding string \citep{Vaswani2017,radford2019language}.\footnote{For space reasons, we do not give a detailed description of the transformer architecture. See \cref{sec:implementation-transformer-models} for the details of the models used in the experiments.}
    This can intuitively be connected to the way \ngram LMs compute the next-symbol probabilities by attending to the last $\ngr-1$ symbols.

    \paragraph{Transformers can represent \ngram LMs.}
    \citet[Thms. 3.1 and 3.2]{svete2024transformers} show that, for any \ngram LM, there exists a weakly equivalent transformer LM.
    They describe intuitive mechanisms in which $\ngr - 1$ heads (Thm. 3.1) or $\ngr - 1$ layers (Thm. 3.2) attend to the previous $\ngr - 1$ symbols using \emph{sparse} attention that computes attention weights with the $\sparsemax$ function \citep{sparsemax}.
    The use of sparse attention is particularly noteworthy since it departs from the standard soft-attention transformers usually used in practice \citep{Vaswani2017}.
    While the theory does not consider soft-attention transformers concretely, it does suggest that being able to assign no attention to irrelevant tokens, which is impossible with soft attention, could be beneficial.
    % To the best of our knowledge, this is the first result connecting transformers to a class of probability distributions, meaning that we can evaluate its empirical implications in the framework recently proposed by \citet{borenstein2024languages}, which we introduce next.
    
    \section{Learnability of \ngram LMs} \label{sec:testing}

    At a high level, we study how well transformers can learn \ngram LMs.
    We do that by training transformers on strings sampled from randomly generated \ngram LMs and measuring a notion of distance between the trained and the ground-truth LM.\looseness=-1

    \subsection{A Framework for Evaluating LMs} \label{sec:framework}
    Given an \ngram LM $\pLMn$ and a transformer LM $\tfpLM$, their \defn{KL divergence} is defined as
    % \begin{subequations}
    %     \begin{align}
    %         \KLFun{\pLMn}{\tfpLM} & \defeq \sum_{\str \in \kleene{\alphabet}} \pLMn\left(\str\right) \log\frac{\pLMn\left(\str\right)}{\pLMFun{\str}} \\
    %                             & = \entropy(\pLMn, \tfpLM) - \entropy(\pLMn) \label{eq:entropy-diff}.
    %     \end{align}
    % \end{subequations}
    \begin{equation}
        \KLFun{\pLMn}{\tfpLM} \defeq \sum_{\str \in \kleene{\alphabet}} \pLMn\left(\str\right) \log\frac{\pLMn\left(\str\right)}{\tfpLM\left(\str\right)}.
    \end{equation}
    The KL divergence is an established measure of similarity between probability distributions.\footnote{KL divergence is not a distance, as it is not symmetric and does not fulfill the triangle inequality.}
    As such, it lends itself naturally to measuring the difference between LMs; in our case, measuring how well a neural LM matches an \ngram LM.\footnote{$\KL$ is also the quantity minimized when training LMs under the MLE objective \citep{cotterell2024formalaspectslanguagemodeling}.}
    This sidesteps the reliance on proxy measures of fit such as the next-symbol prediction accuracy, which is often used for evaluating the learnability of formal languages \citep{borenstein2024languages}.
    We evaluate model performance by approximating $\KL$ on held-out sets; see \cref{sec:evaluation} for details.

    \subsection{Experimental Setup} \label{sec:experimental-setup}
    We study the learnability of \ngram LMs by 
    \begin{enumerate*}[label=\textit{(\arabic*)}]
        \item sampling \ngram LMs of three types described in \cref{sec:data-generation},
        \item generating train\footnote{The development dataset which we use to select the hyperparameters is taken from the training dataset.} and test datasets,
        \item training baselines and transformers, and
        \item computing the KL divergence between the trained LMs and the ground-truth LMs.
    \end{enumerate*}

    \paragraph{Axes of comparison.}
    We investigate the learnability of \ngram LMs along two axes:
    \begin{enumerate}[nosep,label=\textit{Axis \arabic*:},left=0pt]
        \item Whether the \ngram LM is representation-based or not and the degree to which its parameters are shared, and
        \item Measures of complexity of the \ngram LM:
        \begin{itemize}[nosep,left=-1.5em]
            \item the order $\ngr$ of the \ngram LM,
            \item the size $\nsymbols$ of the alphabet, and
            \item the rank $\rank$ of the output matrix $\outMtx$.
        \end{itemize}
    \end{enumerate}
    The first axis is motivated by the parameter-sharing nature of transformers while the second is motivated by the wish to understand the effect of the complexity of the \ngram LM on its learnability.

    \subsubsection{Data Generation} \label{sec:data-generation}
    We generate training and test datasets by sampling strings from randomly generated \ngram LMs of three types: 
    \begin{enumerate*}[label=\textit{(\arabic*)}]
        \item non-representation-based, general, \ngram LMs whose conditional next-symbol probability distributions are arbitrary,
        \item sparse representation-based \ngram LMs, and
        \item dense representation-based \ngram LMs.
    \end{enumerate*}
    The difference between sparse and dense representation-based \ngram LMs lies in the degree to which the parameters of the \ngram LM are shared---in the former, individual symbols of the alphabet define independent parameters (meaning that the size of the representations grows with $\nsymbols$), whereas, in the latter, the parameters are shared across symbols (meaning that the size of the representations is independent of $\nsymbols$).
    See \cref{sec:lm-sampling} for more details.

    We control the complexity of the \ngram LM through the following parameters: $\ngr$, $\nsymbols$, and, for representation-based LMs, the rank $\rank$ of the output matrix $\outMtx$.\footnote{The rank $\rank$ has been identified as a significant predictor of the learnability of general LMs \citep{borenstein2024languages}.}
    Concretely, we generate representation-based datasets with $\ngr \in \set{4, 8, 12}$ and $\nsymbols \in \set{64, 128, 256}$.
    For dense representation-based \ngram LMs, we vary $\outMtx$'s rank $\rank \in \set{2, 8, 16}$.
    Due to the space complexity of storing $\nsymbols^{\ngr - 1}$ different conditional distributions, we generate general \ngram LMs with $\ngr \in \set{2, 4, 6}$ and $\nsymbols = \set{8, 12, 16}$ and additional dense representation-based \ngram LMs of the same complexity for comparison.
    We generate five random \ngram LMs for each of the configurations and sample disjoint training and test datasets of size $\datasetSizeTrain = 50k$ and $\datasetSizeTest = 30k$, respectively.
    % This results in $\left(3 \cdot 3 + 3 \cdot 3 \cdot 3 + 3 \cdot 3 + 3 \cdot 3\right) \cdot 5 = 270$ distinct \ngram LMs.
    See \cref{sec:lm-sampling} for more details.

    \subsubsection{Models}
    % We train classic \ngram estimation techniques and transformers on the generated datasets.

    \paragraph{Transformer models.}
    We use transformers (TF) of different sizes.
    Inspired by the theoretical constructions considered, we are particularly interested in how the number of attention heads and layers affects the learnability of \ngram LMs.
    We investigate this by varying the number of heads and layers.
    Moreover, since the theoretical constructions rely on \emph{sparse} attention mechanisms, we also investigate the effect of using the sparsemax \citep{sparsemax} for computing attention weights.
    More details are given in \cref{sec:learning}.

    \begin{table}
        \centering
        \begin{tabular}{crp{2.75cm}}
            \toprule
                    & \textbf{Parameter}                & \textbf{Description}         \\
            \midrule
            $\pLMn$  & $\ngr \in \N$                       & Order of $\pLMn$             \\
            $\pLMn$  & $\nsymbols \in \N$                  & Size of $\alphabet$          \\
            $\pLMn$  & $\entropy\left(\pLMn\right) \in \R$ & Entropy of $\pLM$            \\
            $\pLMn$  & $\rank \in \N$ & Rank of $\outMtx$            \\
            $\pLMn$  & Dense $\in \set{0, 1}$ & Dense representations            \\
            % $\pLMn$  & $\E\left[|\str|\right]$      & Expected length under $\pLM$ \\
            % $\pLMn$  & $\sfrac{\numOf{\history}}{\nsymbols^{\ngr - 1}}$      & Proportion of seen histories \\
            \midrule
            $\tfpLM$ & $\tfnumlayer \in \N$                & Number of layers             \\
            $\tfpLM$ & $\tfheadnum \in \N$                 & Number of heads              \\
            $\tfpLM$ & $\sparsemax \in \set{0, 1}$                 & Sparse attention              \\
            \bottomrule
        \end{tabular}
        \caption{Predictors of $\KLFun{\pLMn}{\tfpLM}$.}
        \label{tab:predictors}
    \end{table}
    
    \paragraph{Classic techniques.}
    Count-based estimators such as the maximum-likelihood \ngram estimate have been used in NLP for decades.
    They compute the next-symbol probabilities by counting the occurrences of (lower-order) \ngram{}s in the training data.
    They are thus well-suited for learning \ngram LMs, making them a difficult-to-beat baseline.
    \emph{Smoothing} estimates the probabilities of unseen \ngram{}s by redistributing the probability mass of seen \ngram{}s, regularizing count-based estimators \citep{katz,witten_bell,absolute_discounting,simpleGT,malagutti2024role}.\footnote{We expect that the fact that the ground-truth \ngram LMs are of a high order and representation-based makes the task for traditional smoothing techniques more difficult and may require a large degree of smoothing.}
    We consider add-$\lambda$, absolute discounting \citep{absolute_discounting}, and Witten--Bell \citep{witten_bell} smoothing along with the standard maximum likelihood solution. 
    These techniques are described in \cref{sec:estimation-techniques}; see also \cref{sec:learning} for details.

    \begin{table*}
        \centering
        \begin{tabular}{lcccccc}
            \toprule
            $\nsymbols$ & \multicolumn{2}{c}{$8$} & \multicolumn{2}{c}{$12$} & \multicolumn{2}{c}{$16$} \\
            \cmidrule(lr){2-3} \cmidrule(lr){4-5} \cmidrule(lr){6-7} Parameter sharing & No & Yes & No & Yes & No & Yes \\
            \midrule
            \textbf{Classic} & \meanStd{\mathbf{3.04}}{1.46} & \meanStd{2.36}{0.23} & \meanStd{\mathbf{17.34}}{1.01} & \meanStd{2.45}{0.96} & \meanStd{\mathbf{50.35}}{1.87} & \meanStd{3.11}{0.42} \\
            \textbf{LL}  & \meanStd{101.42}{3.46} & \meanStd{21.96}{3.38} & \meanStd{109.60}{2.06} & \meanStd{27.10}{4.95} & \meanStd{117.83}{2.37} & \meanStd{28.37}{3.53} \\
            \textbf{Neural}  & \meanStd{60.87}{2.87} & \meanStd{\mathbf{1.50}}{0.20} & \meanStd{79.77}{1.63} & \meanStd{\mathbf{1.43}}{0.78} & \meanStd{90.01}{1.18} & \meanStd{\mathbf{1.63}}{0.47} \\
            \midrule
            \textbf{TF}   & \meanStd{67.06}{2.91} & \meanStd{4.63}{2.00} & \meanStd{77.95}{1.74} & \meanStd{2.80}{0.72} & \meanStd{86.38}{1.88} & \meanStd{3.68}{1.33} \\
            \bottomrule
        \end{tabular}
        \caption{Learnability of general and dense representation-based \ngram LMs for $\ngr = 6$.}
        \label{tab:general-vs-representations-results}
    \end{table*}
    
    \paragraph{Two additional baselines.}
    We study two additional baselines: a \defn{log-linear model} (LL) with fixed, sparse representations of the histories and a \defn{neural \ngram LM} (Neural) that learns dense representations of the histories.
    The log-linear model represents the history $\nstr$ as a concatenation of the one-hot encodings of its symbols.
    It learns the appropriate output matrix $\widehat{\outMtx} \in \R^{\eosnsymbols \times \left(\ngr - 1\right) \bosnsymbols}$ such that $\pLMn(\eossym\mid \nstr) = \softmax(\widehat{\outMtx} \; \enc\left(\nstr\right))_{\eossym}$ approximates the training data well.
    The neural \ngram LM is based on previous work exploring neural \ngram LMs \citep{NIPS2000_728f206c,10.5555/944919.944966,Bengio2006,sun-iyyer-2021-revisiting}.
    It learns static symbol representations $\repFuncFun{\sym} \in \R^D$ for $\sym \in \alphabet$ and concatenates $\repFuncFun{\sym_i}$ for $\nstr = \sym_1 \cdots \sym_{\ngr - 1}$ before non-linearly transforming the concatenated representations with an MLP to compute  $\enc\left(\nstr\right)$.
    See also \cref{sec:estimation-techniques}.

    \subsubsection{Statistical Analysis}
    We determine the importance of \ngram LM parameters and transformer components on the learnability of \ngram LMs by evaluating how predictive they are of the transformer performance.
    Concretely, we investigate the predictors listed in \cref{tab:predictors} and fit a linear regression model to predict the test KL divergence.
    See \cref{sec:stat-analysis} for more details.

    \begin{table*}
        \centering
        \begin{tabular}{lccccccccc}
            \toprule
            $\ngr$           & \multicolumn{3}{c}{$4$}          & \multicolumn{3}{c}{$8$} & \multicolumn{3}{c}{$12$}                                                                                                                                                                                                                   \\
            \cmidrule(lr){2-4} \cmidrule(lr){5-7} \cmidrule(lr){8-10}
            $\nsymbols$ & $64$ & $128$ & $256$ & $64$ & $128$ & $256$ & $64$ & $128$ & $256$ \\
            \midrule
            % \textbf{MLE}           & \meanStd{3.25}{0.58} & \meanStd{6.57}{1.29} & \meanStd{12.05}{1.86} & \meanStd{73.70}{41.01} & \meanStd{235.19}{32.65} & \meanStd{298.84}{30.01} & \meanStd{245.39}{57.65} & \meanStd{300.87}{26.61} & \meanStd{340.88}{33.70}         \\
            \textbf{Classic}            & \meanStd{2.11}{0.39} & \meanStd{3.40}{0.81} & \meanStd{5.00}{0.72} & \meanStd{25.72}{8.48} & \meanStd{54.95}{4.17} & \meanStd{67.64}{5.25} & \meanStd{58.17}{9.36} & \meanStd{74.09}{6.78} & \meanStd{89.90}{6.96}  \\
            \textbf{LL} & \meanStd{32.05}{3.14} & \meanStd{37.10}{9.53} & \meanStd{40.61}{3.93} & \meanStd{40.45}{11.01} & \meanStd{58.26}{3.64} & \meanStd{62.00}{5.23} & \meanStd{43.40}{5.46} & \meanStd{49.75}{4.12} & \meanStd{61.60}{5.45}
            \\
            \textbf{Neural}            & \meanStd{\mathbf{1.14}}{0.64} & \meanStd{\mathbf{1.98}}{0.91} & \meanStd{\mathbf{2.17}}{1.46} & \meanStd{\mathbf{5.96}}{1.86} & \meanStd{\mathbf{9.13}}{0.88} & \meanStd{\mathbf{9.06}}{0.68} & \meanStd{\mathbf{7.71}}{0.75} & \meanStd{\mathbf{9.87}}{1.28} & \meanStd{\mathbf{11.20}}{1.09}  \\
            \midrule
            \textbf{TF}             & \meanStd{2.43}{0.51} & \meanStd{5.04}{1.75} & \meanStd{5.63}{1.80} & \meanStd{9.52}{1.98} & \meanStd{14.72}{0.95} & \meanStd{16.89}{1.33} & \meanStd{14.73}{1.70} & \meanStd{22.79}{6.42} & \meanStd{33.99}{9.06}  \\
            \bottomrule
        \end{tabular}
        \caption{The effect of the order $\ngr$ and the alphabet size $\nsymbols$ on the performance of the best performing models.}
        \label{tab:best-results}
    \end{table*}

    \begin{table}
        \centering
        \fontsize{10}{10}\selectfont
        % \sisetup{table-format=3.2, group-minimum-digits=3}
        \begin{tabular}{lcc}
            \toprule
            Predictor                    & $\widehat{\beta}$  & $p$-value \\
            \midrule
            Intercept                    &  \meanStdSameRow{0.99}{0.03}                                     &   $< 10^{-6}$           \\
            \midrule
            $\ngr$                       &   \meanStdSameRow{\textcolor{ETHRed}{0.63}}{0.01}             &   $< 10^{-6}$    \\
            $\nsymbols$                  &   \meanStdSameRow{\textcolor{ETHRed}{0.28}}{0.01}             &   $< 10^{-6}$       \\
            $\rank$                      &  \meanStdSameRow{\textcolor{ETHRed}{0.17}}{0.02}              &   $< 10^{-6}$       \\
            $\entropy\left(\pLMn\right)$ &   \meanStdSameRow{\textcolor{ETHGreen}{-0.05}}{0.02}             &   $0.034$     \\
            Dense                        &   \meanStdSameRow{\textcolor{ETHGreen}{-1.28}}{0.04}             &   $< 10^{-6}$     \\
            \midrule
            $\tfnumlayer$                &   \meanStdSameRow{\textcolor{ETHGreen}{-0.33}}{0.01}             &   $< 10^{-6}$          \\
            $\tfheadnum$                 &   \meanStdSameRow{\textcolor{ETHGreen}{-0.15}}{0.01}             &   $< 10^{-6}$       \\
            $\sparsemax$                 &   \meanStdSameRow{\textcolor{ETHGreen}{-0.06}}{0.02}             &   $0.002$      \\
            \bottomrule
        \end{tabular}
        \caption{Estimated coefficients ($\widehat{\beta}$) and $p$-values of the linear model predicting $\widehat{\KL}$. The $R^2$ value is $0.665$.}
        \label{tab:tf-coefficients}
        \vspace{-2.5pt}
    \end{table}

    \section{Experimental Results and Discussion} \label{sec:results}
    This section presents the experimental results.
    All tables contain the mean and standard deviation of the estimate of the KL divergence $\widehat{\KL}$ between the ground truth and trained LMs estimated on the test dataset.\footnote{See \cref{sec:evaluation} for details on the approximation.}
    % \textbf{Bolded} results indicate the best-performing model in the given setting (column).
    Negative values indicate an approximation error of the (low) $\KL$.
    
    \subsection{The Effect of Parameter Sharing} \label{sec:results-parameter-sharing}
    % We first analyze the effect of whether the \ngram LM is representation-based or not.
    \cref{tab:general-vs-representations-results} compares the performance of transformers and baselines on representation- and non-representation-based \ngram LMs with $\ngr = 6$.
    As expected, simple counting-based methods that do not assume representation-based ground-truth LMs perform equally well regardless of whether or not the \ngram LM is representation-based.
    % well on representation- and non-representation-based \ngram LMs.
    Models that assume representation-based \ngram LMs (log-linear, neural \ngram, and transformer LMs) perform better on representation-based ones.
    Transformers outperform log-linear models, possibly due to their better fit for dense representation-based models, and the simple neural \ngram LMs perform best.
    \cref{sec:additional-results-onehot-vs-vectorial} further analyzes the effect of the degree to which the parameters of the \ngram LM are shared, showing that sparse-representation-based \ngram LMs are more difficult to approximate with the neural \ngram model and transformers than dense-representation-based ones.

    Our results provide a complementary view of the theory connecting transformers to \ngram LMs.
    They show the importance of \emph{parameter sharing} in the ground-truth model for transformers to match the distribution well---in the case of no parameter sharing, simple count-based methods perform better.
    % \footnote{Although the rank of $\outMtx$ was not found to be a significant predictor of the performance, the fact that transformers do significantly better on dense-representation-based \ngram LMs than on sparse-representation-based ones provides further evidence for the importance of the LM rank, as reported by \citet{borenstein2024languages}}.
    The good performance of transformers is particularly interesting compared to the simple and more structured log-linear model.\looseness=-1
    
    \subsection{The effect of \ngram LM Complexity} \label{sec:results-complexity}
    We next investigate the effect of the \ngram LM parameters described in \cref{sec:experimental-setup}.
    \cref{tab:best-results} show the performance of the best model configurations for different combinations of $\ngr$ and $\nsymbols$ on \ngram LMs with dense representations and of rank $16$. 
    As expected, the performance of all tested methods decays with increasing \ngram complexity.
    While the structured neural \ngram models perform best, transformers do not fare much worse, outperforming classical methods on all but the smallest models and the log-linear model in all settings.
    Rank is further investigated in \cref{sec:rank-results}, where we show that higher-rank models are more difficult to learn for neural \ngram and transformer LMs.

    \subsection{Predictors of Learnability}
    \cref{tab:tf-coefficients} shows the linear model coefficients predicting the transformer performance.
    According to the absolute values of the coefficients, the most significant predictor is whether the representations are dense or sparse, followed by the \ngram LM order and the alphabet size.
    The entropy is \emph{negatively} correlated with the KL divergence, indicating that higher-entropy \ngram LMs are easier to learn, in line with existing work \citep{borenstein2024languages}.
    As suggested by theoretical work, we find the number of heads and layers to be significant positive predictors of the learnability of the \ngram LM.

    \paragraph{Soft vs. sparse attention transformers.}
    The distinction between soft- and sparse-attention transformers made by \citet{svete2024transformers} (and the distinction between soft- and hard-attention transformers in theoretical literature) makes the impact of using sparse attention particularly interesting.
    As shown by \cref{tab:tf-coefficients}, using sparse attention is a significant predictor of \emph{better} performance, which agrees with the intuition that zero attention weights are useful for representing \ngram LMs. 
    This is further investigated in \cref{sec:sparse-attention-results}, showing the superior performance of sparse-attention transformers.
    While these results agree with the intuition regarding the importance of sparse attention made by \citet{svete2024transformers}, they depart from the softmax-computed attention weights normally used in practice \citep{Vaswani2017}.
    This encourages further investigation into the possible benefits of sparse attention mechanisms in transformers, particularly in the context of learning formal languages---could sparse attention, for example, aid generalization in situations where non-zero attention weights could lead to accumulating errors, such as those in counting \citep{weiss-etal-2018-practical}?

    \section{Conclusion}

    We compare the performance of transformers to that of classical \ngram estimation techniques and hand-crafted baselines at learning artificially generated \ngram LMs.
    Transformers show a good inductive bias towards learning representation-based \ngram LMs but fare worse than classic estimation techniques at learning general \ngram LMs, underlining the importance of parameter sharing in the ground-truth model for transformers to learn it well.
    Moreover, the impact of the number of heads and layers on the performance agrees with existing theoretical results.
    The better performance of sparse-attention transformers motivates further investigation into how sparse attention could be used when learning formal languages with transformers.

    \section*{Limitations}

    We highlight some limitations of our work.
    First, note that the concrete theoretical results this paper investigates rely on a formulation of the transformer architecture that is not practical for training---most notably, \citeposs{svete2024transformers} results either assume the use of hard attention or the use of the more practical sparse attention but unbounded positional encodings, which are not commonly used in practice.
    We replace these modeling assumptions with more practical ones---we use the softmax and sparsemax normalization functions and bounded positional encodings.
    We also note that the LMs we use to evaluate transformers---\ngram LMs---are very simple, which makes the results less generalizable to more complex LMs.
    This is done on purpose to stay as close as possible to the theoretical results connecting transformers to \ngram LMs.

    \section*{Acknowledgements}
    We thank Luca Malagutti for his help during the early stages of this project.
    Ryan Cotterell acknowledges support from the Swiss National Science
    Foundation (SNSF) as part of the ``The Forgotten Role of Inductive Bias in Interpretability'' project.
    Anej Svete is supported by the ETH AI Center Doctoral Fellowship.
    This research was partially funded by a DFF Sapere Aude research leader grant under grant agreement No 0171-00034B, the Danish--Israeli Study Foundation in Memory of Josef and Regine Nachemsohn, and the Privacy Black \& White project, a UCPH Data+ Grant. This work was further supported by the Pioneer Centre for AI, DNRF grant number P1.

    \bibliography{anthology,custom}

\begin{thebibliography}{39}
\expandafter\ifx\csname natexlab\endcsname\relax\def\natexlab#1{#1}\fi

\bibitem[{Akyürek et~al.(2024)Akyürek, Wang, Kim, and Andreas}]{akyürek2024incontext}
Ekin Akyürek, Bailin Wang, Yoon Kim, and Jacob Andreas. 2024.
\newblock \href {http://arxiv.org/abs/2401.12973} {In-context language learning: Architectures and algorithms}.
\newblock \emph{arXiv preprint arXiv:2401.12973}.

\bibitem[{Arora et~al.(2024)Arora, Eyuboglu, Timalsina, Johnson, Poli, Zou, Rudra, and R{\'{e}}}]{DBLP:conf/iclr/AroraETJP0RR24}
Simran Arora, Sabri Eyuboglu, Aman Timalsina, Isys Johnson, Michael Poli, James Zou, Atri Rudra, and Christopher R{\'{e}}. 2024.
\newblock \href {https://openreview.net/forum?id=LY3ukUANko} {Zoology: Measuring and improving recall in efficient language models}.
\newblock In \emph{The Twelfth International Conference on Learning Representations, {ICLR} 2024, Vienna, Austria, May 7-11, 2024}. OpenReview.net.

\bibitem[{Bengio et~al.(2000)Bengio, Ducharme, and Vincent}]{NIPS2000_728f206c}
Yoshua Bengio, R\'{e}jean Ducharme, and Pascal Vincent. 2000.
\newblock \href {https://proceedings.neurips.cc/paper_files/paper/2000/file/728f206c2a01bf572b5940d7d9a8fa4c-Paper.pdf} {A neural probabilistic language model}.
\newblock In \emph{Advances in Neural Information Processing Systems}, volume~13. MIT Press.

\bibitem[{Bengio et~al.(2003)Bengio, Ducharme, Vincent, and Janvin}]{10.5555/944919.944966}
Yoshua Bengio, R\'{e}jean Ducharme, Pascal Vincent, and Christian Janvin. 2003.
\newblock \href {https://www.jmlr.org/papers/volume3/bengio03a/bengio03a.pdf} {A neural probabilistic language model}.
\newblock \emph{J. Mach. Learn. Res.}, 3:1137–1155.

\bibitem[{Bengio et~al.(2006)Bengio, Schwenk, Sen{\'e}cal, Morin, and Gauvain}]{Bengio2006}
Yoshua Bengio, Holger Schwenk, Jean-S{\'e}bastien Sen{\'e}cal, Fr{\'e}deric Morin, and Jean-Luc Gauvain. 2006.
\newblock \href {https://doi.org/10.1007/3-540-33486-6_6} {\emph{Neural Probabilistic Language Models}}, pages 137--186. Springer Berlin Heidelberg, Berlin, Heidelberg.

\bibitem[{Bhattamishra et~al.(2020)Bhattamishra, Ahuja, and Goyal}]{bhattamishra-etal-2020-on-ability}
Satwik Bhattamishra, Kabir Ahuja, and Navin Goyal. 2020.
\newblock \href {https://doi.org/10.18653/v1/2020.emnlp-main.576} {On the ability and limitations of transformers to recognize formal languages}.
\newblock In \emph{Proceedings of the 2020 Conference on Empirical Methods in Natural Language Processing (EMNLP)}, pages 7096--7116, Online. Association for Computational Linguistics.

\bibitem[{Bird et~al.(2009)Bird, Klein, and Loper}]{bird2009natural}
Steven Bird, Ewan Klein, and Edward Loper. 2009.
\newblock \href {https://www.nltk.org/book/} {\emph{Natural language processing with Python: analyzing text with the natural language toolkit}}.
\newblock " O'Reilly Media, Inc.".

\bibitem[{Borenstein et~al.(2024)Borenstein, Svete, Chan, Valvoda, Nowak, Augenstein, Chodroff, and Cotterell}]{borenstein2024languages}
Nadav Borenstein, Anej Svete, Robin Chan, Josef Valvoda, Franz Nowak, Isabelle Augenstein, Eleanor Chodroff, and Ryan Cotterell. 2024.
\newblock \href {http://arxiv.org/abs/2406.04289} {What languages are easy to language-model? a perspective from learning probabilistic regular languages}.
\newblock \emph{arXiv preprint arXiv:2406.04289}.

\bibitem[{Chen et~al.(2024)Chen, Shwartz{-}Ziv, Cho, Leavitt, and Saphra}]{DBLP:conf/iclr/ChenSCLS24}
Angelica Chen, Ravid Shwartz{-}Ziv, Kyunghyun Cho, Matthew~L. Leavitt, and Naomi Saphra. 2024.
\newblock \href {https://openreview.net/forum?id=MO5PiKHELW} {Sudden drops in the loss: Syntax acquisition, phase transitions, and simplicity bias in mlms}.
\newblock In \emph{The Twelfth International Conference on Learning Representations, {ICLR} 2024, Vienna, Austria, May 7-11, 2024}. OpenReview.net.

\bibitem[{Chen and Goodman(1999)}]{chen_empirical}
Stanley~F. Chen and Joshua Goodman. 1999.
\newblock \href {https://doi.org/https://doi.org/10.1006/csla.1999.0128} {An empirical study of smoothing techniques for language modeling}.
\newblock \emph{Computer Speech \& Language}, 13(4):359--394.

\bibitem[{Cotterell et~al.(2024)Cotterell, Svete, Meister, Liu, and Du}]{cotterell2024formalaspectslanguagemodeling}
Ryan Cotterell, Anej Svete, Clara Meister, Tianyu Liu, and Li~Du. 2024.
\newblock \href {http://arxiv.org/abs/2311.04329} {Formal aspects of language modeling}.
\newblock \emph{arXiv preprint arXiv:2311.04329}.

\bibitem[{Del{\'{e}}tang et~al.(2023)Del{\'{e}}tang, Ruoss, Grau{-}Moya, Genewein, Wenliang, Catt, Cundy, Hutter, Legg, Veness, and Ortega}]{deletang2023neural}
Gr{\'{e}}goire Del{\'{e}}tang, Anian Ruoss, Jordi Grau{-}Moya, Tim Genewein, Li~Kevin Wenliang, Elliot Catt, Chris Cundy, Marcus Hutter, Shane Legg, Joel Veness, and Pedro~A. Ortega. 2023.
\newblock \href {https://openreview.net/forum?id=WbxHAzkeQcn} {Neural networks and the {C}homsky hierarchy}.
\newblock In \emph{11th International Conference on Learning Representations}.

\bibitem[{Edelman et~al.(2024)Edelman, Edelman, Goel, Malach, and Tsilivis}]{edelman2024evolutionstatisticalinductionheads}
Benjamin~L. Edelman, Ezra Edelman, Surbhi Goel, Eran Malach, and Nikolaos Tsilivis. 2024.
\newblock \href {http://arxiv.org/abs/2402.11004} {The evolution of statistical induction heads: In-context learning markov chains}.
\newblock \emph{arXiv preprint arXiv:2402.11004}.

\bibitem[{Eisner(2002)}]{eisner-2002-parameter}
Jason Eisner. 2002.
\newblock \href {https://doi.org/10.3115/1073083.1073085} {Parameter estimation for probabilistic finite-state transducers}.
\newblock In \emph{Proceedings of the 40th Annual Meeting of the Association for Computational Linguistics}, pages 1--8, Philadelphia, Pennsylvania, USA. Association for Computational Linguistics.

\bibitem[{Gale and Sampson(1995)}]{simpleGT}
William~A. Gale and Geoffrey Sampson. 1995.
\newblock \href {https://doi.org/10.1080/09296179508590051} {Good‐turing frequency estimation without tears}.
\newblock \emph{Journal of Quantitative Linguistics}, 2(3):217--237.

\bibitem[{Katz(1987)}]{katz}
S.~Katz. 1987.
\newblock \href {https://doi.org/10.1109/TASSP.1987.1165125} {Estimation of probabilities from sparse data for the language model component of a speech recognizer}.
\newblock \emph{IEEE Transactions on Acoustics, Speech, and Signal Processing}, 35(3):400--401.

\bibitem[{Kingma and Ba(2017)}]{kingma2017adam}
Diederik~P. Kingma and Jimmy Ba. 2017.
\newblock \href {http://arxiv.org/abs/1412.6980} {Adam: A method for stochastic optimization}.
\newblock \emph{arXiv preprint arXiv:1412.6980}.

\bibitem[{Liu et~al.(2024)Liu, Min, Zettlemoyer, Choi, and Hajishirzi}]{liu2024infinigramscalingunboundedngram}
Jiacheng Liu, Sewon Min, Luke Zettlemoyer, Yejin Choi, and Hannaneh Hajishirzi. 2024.
\newblock \href {http://arxiv.org/abs/2401.17377} {Infini-gram: Scaling unbounded n-gram language models to a trillion tokens}.
\newblock \emph{arXiv preprint arXiv:2401.17377}.

\bibitem[{Loshchilov and Hutter(2018)}]{loshchilov2018fixing}
Ilya Loshchilov and Frank Hutter. 2018.
\newblock \href {https://openreview.net/forum?id=rk6qdGgCZ} {Fixing weight decay regularization in {Adam}}.
\newblock \emph{arXiv.org}.

\bibitem[{Malagutti et~al.(2024)Malagutti, Buinovskij, Svete, Meister, Amini, and Cotterell}]{malagutti2024role}
Luca Malagutti, Andrius Buinovskij, Anej Svete, Clara Meister, Afra Amini, and Ryan Cotterell. 2024.
\newblock \href {http://arxiv.org/abs/2403.17240} {The role of $n$-gram smoothing in the age of neural networks}.
\newblock \emph{arXiv preprint arXiv:2403.17240}.

\bibitem[{Martins and Astudillo(2016)}]{sparsemax}
Andr\'{e} F.~T. Martins and Ram\'{o}n~F. Astudillo. 2016.
\newblock \href {https://proceedings.mlr.press/v48/martins16} {From softmax to sparsemax: {A} sparse model of attention and multi-label classification}.
\newblock In \emph{Proceedings of the 33rd International Conference on International Conference on Machine Learning - Volume 48}, ICML'16, page 1614–1623.

\bibitem[{Ney et~al.(1994)Ney, Essen, and Kneser}]{absolute_discounting}
Hermann Ney, Ute Essen, and Reinhard Kneser. 1994.
\newblock \href {https://www.sciencedirect.com/science/article/abs/pii/S0885230884710011} {On structuring probabilistic dependences in stochastic language modelling}.
\newblock \emph{Computer Speech \& Language}, 8(1):1--38.

\bibitem[{Nguyen(2024)}]{nguyen2024understandingtransformersngramstatistics}
Timothy Nguyen. 2024.
\newblock \href {http://arxiv.org/abs/2407.12034} {Understanding transformers via n-gram statistics}.
\newblock \emph{arXiv preprint arXiv:2407.12034}.

\bibitem[{Nowak et~al.(2024)Nowak, Svete, Butoi, and Cotterell}]{nowak-etal-2024-computational}
Franz Nowak, Anej Svete, Alexandra Butoi, and Ryan Cotterell. 2024.
\newblock On the representational capacity of neural language models with chain-of-thought reasoning.
\newblock In \emph{Proceedings of the 62nd Annual Meeting of the Association for Computational Linguistics (Volume 1: Long Papers)}, Bangkok, Thailand. Association for Computational Linguistics.

\bibitem[{Nowak et~al.(2023)Nowak, Svete, Du, and Cotterell}]{nowak-etal-2023-representational}
Franz Nowak, Anej Svete, Li~Du, and Ryan Cotterell. 2023.
\newblock \href {https://doi.org/10.18653/v1/2023.emnlp-main.434} {On the representational capacity of recurrent neural language models}.
\newblock In \emph{Proceedings of the 2023 Conference on Empirical Methods in Natural Language Processing}, pages 7011--7034, Singapore. Association for Computational Linguistics.

\bibitem[{Olsson et~al.(2022)Olsson, Elhage, Nanda, Joseph, DasSarma, Henighan, Mann, Askell, Bai, Chen, Conerly, Drain, Ganguli, Hatfield-Dodds, Hernandez, Johnston, Jones, Kernion, Lovitt, Ndousse, Amodei, Brown, Clark, Kaplan, McCandlish, and Olah}]{olsson2022context}
Catherine Olsson, Nelson Elhage, Neel Nanda, Nicholas Joseph, Nova DasSarma, Tom Henighan, Ben Mann, Amanda Askell, Yuntao Bai, Anna Chen, Tom Conerly, Dawn Drain, Deep Ganguli, Zac Hatfield-Dodds, Danny Hernandez, Scott Johnston, Andy Jones, Jackson Kernion, Liane Lovitt, Kamal Ndousse, Dario Amodei, Tom Brown, Jack Clark, Jared Kaplan, Sam McCandlish, and Chris Olah. 2022.
\newblock \href {https://transformer-circuits.pub/2022/in-context-learning-and-induction-heads/index.html} {In-context learning and induction heads}.
\newblock \emph{Transformer Circuits Thread}.

\bibitem[{Paszke et~al.(2019)Paszke, Gross, Massa, Lerer, Bradbury, Chanan, Killeen, Lin, Gimelshein, Antiga, Desmaison, Köpf, Yang, DeVito, Raison, Tejani, Chilamkurthy, Steiner, Fang, Bai, and Chintala}]{paszke2019pytorch}
Adam Paszke, Sam Gross, Francisco Massa, Adam Lerer, James Bradbury, Gregory Chanan, Trevor Killeen, Zeming Lin, Natalia Gimelshein, Luca Antiga, Alban Desmaison, Andreas Köpf, Edward Yang, Zach DeVito, Martin Raison, Alykhan Tejani, Sasank Chilamkurthy, Benoit Steiner, Lu~Fang, Junjie Bai, and Soumith Chintala. 2019.
\newblock \href {http://arxiv.org/abs/1912.01703} {Pytorch: An imperative style, high-performance deep learning library}.
\newblock \emph{arXiv preprint arXiv:1912.01703}.

\bibitem[{Radford et~al.(2019)Radford, Wu, Child, Luan, Amodei, and Sutskever}]{radford2019language}
Alec Radford, Jeff Wu, Rewon Child, D.~Luan, Dario Amodei, and Ilya Sutskever. 2019.
\newblock \href {https://www.semanticscholar.org/paper/Language-Models-are-Unsupervised-Multitask-Learners-Radford-Wu/9405cc0d6169988371b2755e573cc28650d14dfe} {Language models are unsupervised multitask learners}.

\bibitem[{Strobl et~al.(2023)Strobl, Merrill, Weiss, Chiang, and Angluin}]{strobl2023transformers}
Lena Strobl, William Merrill, Gail Weiss, David Chiang, and Dana Angluin. 2023.
\newblock \href {http://arxiv.org/abs/2311.00208} {Transformers as recognizers of formal languages: A survey on expressivity}.
\newblock \emph{arXiv preprint arXiv:2311.00208}.

\bibitem[{Sun and Iyyer(2021)}]{sun-iyyer-2021-revisiting}
Simeng Sun and Mohit Iyyer. 2021.
\newblock \href {https://doi.org/10.18653/v1/2021.naacl-main.407} {Revisiting simple neural probabilistic language models}.
\newblock In \emph{Proceedings of the 2021 Conference of the North American Chapter of the Association for Computational Linguistics: Human Language Technologies}, pages 5181--5188, Online. Association for Computational Linguistics.

\bibitem[{Svete et~al.(2024{\natexlab{a}})Svete, Chan, and Cotterell}]{svete2024theoretical}
Anej Svete, Robin Shing~Moon Chan, and Ryan Cotterell. 2024{\natexlab{a}}.
\newblock \href {http://arxiv.org/abs/2402.15814} {A theoretical result on the inductive bias of {RNN} language models}.
\newblock \emph{arXiv preprint arXiv:2402.15814}.

\bibitem[{Svete and Cotterell(2023)}]{svete-cotterell-2023-recurrent}
Anej Svete and Ryan Cotterell. 2023.
\newblock \href {https://doi.org/10.18653/v1/2023.emnlp-main.502} {Recurrent neural language models as probabilistic finite-state automata}.
\newblock In \emph{Proceedings of the 2023 Conference on Empirical Methods in Natural Language Processing}, pages 8069--8086, Singapore. Association for Computational Linguistics.

\bibitem[{Svete and Cotterell(2024)}]{svete2024transformers}
Anej Svete and Ryan Cotterell. 2024.
\newblock \href {http://arxiv.org/abs/2404.14994} {Transformers can represent $n$-gram language models}.
\newblock \emph{arXiv preprint arXiv:2404.14994}.

\bibitem[{Svete et~al.(2024{\natexlab{b}})Svete, Nowak, Sahabdeen, and Cotterell}]{svete-etal-2024-lower}
Anej Svete, Franz Nowak, Anisha Sahabdeen, and Ryan Cotterell. 2024{\natexlab{b}}.
\newblock \href {https://doi.org/10.18653/v1/2024.naacl-long.380} {Lower bounds on the expressivity of recurrent neural language models}.
\newblock In \emph{Proceedings of the 2024 Conference of the North American Chapter of the Association for Computational Linguistics: Human Language Technologies (Volume 1: Long Papers)}, pages 6820--6844, Mexico City, Mexico. Association for Computational Linguistics.

\bibitem[{Vaswani et~al.(2017)Vaswani, Shazeer, Parmar, Uszkoreit, Jones, Gomez, Kaiser, and Polosukhin}]{Vaswani2017}
Ashish Vaswani, Noam Shazeer, Niki Parmar, Jakob Uszkoreit, Llion Jones, Aidan~N. Gomez, {\L}ukasz Kaiser, and Illia Polosukhin. 2017.
\newblock \href {https://proceedings.neurips.cc/paper_files/paper/2017/file/3f5ee243547dee91fbd053c1c4a845aa-Paper.pdf} {Attention is all you need}.
\newblock In \emph{Advances in Neural Information Processing Systems}, volume~30. Curran Associates, Inc.

\bibitem[{Voita et~al.(2024)Voita, Ferrando, and Nalmpantis}]{voita-etal-2024-neurons}
Elena Voita, Javier Ferrando, and Christoforos Nalmpantis. 2024.
\newblock \href {https://doi.org/10.18653/v1/2024.findings-acl.75} {Neurons in large language models: Dead, n-gram, positional}.
\newblock In \emph{Findings of the Association for Computational Linguistics ACL 2024}, pages 1288--1301, Bangkok, Thailand and virtual meeting. Association for Computational Linguistics.

\bibitem[{Weiss et~al.(2018)Weiss, Goldberg, and Yahav}]{weiss-etal-2018-practical}
Gail Weiss, Yoav Goldberg, and Eran Yahav. 2018.
\newblock \href {https://doi.org/10.18653/v1/P18-2117} {On the practical computational power of finite precision {RNN}s for language recognition}.
\newblock In \emph{Proceedings of the 56th Annual Meeting of the Association for Computational Linguistics (Volume 2: Short Papers)}, pages 740--745, Melbourne, Australia. Association for Computational Linguistics.

\bibitem[{Witten and Bell(1991)}]{witten_bell}
Ian~H. Witten and Timothy~C. Bell. 1991.
\newblock \href {https://doi.org/10.1109/18.87000} {The zero-frequency problem: Estimating the probabilities of novel events in adaptive text compression}.
\newblock \emph{IEEE Transactions on Information Theory}, 37(4):1084--1094.

\bibitem[{Zmigrod et~al.(2021)Zmigrod, Vieira, and Cotterell}]{zmigrod-etal-2021-efficient-computation}
Ran Zmigrod, Tim Vieira, and Ryan Cotterell. 2021.
\newblock \href {https://doi.org/10.1162/tacl_a_00391} {Efficient computation of expectations under spanning tree distributions}.
\newblock \emph{Transactions of the Association for Computational Linguistics}, 9:675--690.

\end{thebibliography}
    \bibliographystyle{acl_natbib}

    \newpage
    \onecolumn

    \appendix

    \section{Classic \ngram Estimation Techniques} \label{sec:estimation-techniques}
    \subsection{Maximum Likelihood Estimation (MLE)}
    The maximum likelihood \ngram LM estimate of order $\ngr$ computes the next-symbol probabilities
    \begin{equation}
        \qngram(\sym\mid\nstr) \defeq \frac{\CountFun{\nstr\sym}}{\CountFun{\nstr}},
    \end{equation}
    for $\sym \in \eosalphabet$ and $\nstr \in \alphabet^{\ngr - 1}$, which counts the total number of times the string $\nstr\sym \in \alphabet^\ngr$ occurs and divides by the number of total occurrences of the history $\nstr \in \alphabet^{\ngr - 1}$. 
    MLE models are prone to overfitting, especially when training data is sparse or if history lengths are long, thus demonstrating the need for smoothing.

    \subsection{Add-$\lambda$ smoothing (Add-$\lambda$)}
    Add-$\lambda$ smoothing is one of the simplest methods of obtaining a smoothed \ngram model from the MLE estimate $\qngram$. 
    Add-$\lambda$ smoothing adds a pseudo-count of $\lambda$ to the counts of all possible \ngram{}s, including those not seen in the training dataset.
    % Given a parameter $\lambda$, we pretend as though each word was counted $\lambda$ more times than previously before, augmenting the counts of all \ngram{}s, including those not previously seen in training, by $\lambda$. 
    Mathematically,
    \begin{equation}
        \CountAS(\nstr\sym) \defeq \CountFun{\nstr\sym}+\lambda
    \end{equation}
    and
    \begin{equation}
        \pAS(\sym\mid\nstr) \defeq \frac{\CountFun{\nstr\sym}+\lambda}{\CountFun{\nstr}+|\Sigma + 1|\lambda}
    \end{equation}
    for $\sym \in \eosalphabet$ and $\nstr \in \alphabet^{\ngr - 1}$.
    In the special case of $\lambda = 1$, we have \textbf{Laplace Smoothing}.

    \subsection{Absolute Discounting (AD) \texorpdfstring{(\citeyear{absolute_discounting})}{}} 
    Absolute Discounting (AD) involves the interpolation of higher and lower-order \ngram models. Though higher-order \ngram{}s capture more context, they often suffer from zero probabilities and overfitting due to limited training data. AD makes higher-order distributions by subtracting a fixed discount $\delta \leq 1$ from each nonzero count and recursively interpolates this with \ngram of lower degree. That is:
    \begin{equation}
        \pAD(\sym\mid\nstr) = \frac{\max\{ \CountFun{\nstr\sym} - \delta, 0\}}{\CountFun{\nstr}} +  (1-\lambda_n) \pADminus(\sym\mid\nmstr)
    \end{equation}
    for $\sym \in \eosalphabet$ and $\nstr \in \alphabet^{\ngr - 1}$.
    To make the probabilities sum to 1, we let
    \begin{equation}
        1 - \lambda_n = \frac{\delta\typecount(\nstr, \bullet)}{\CountFun{\nstr}},
    \end{equation}
    where $\typecount$ is the \emph{type counts} function of order $n$, formally defined as
    \begin{equation}
        \typecount(\nstr, \sym) \defeq \begin{cases}
            1 & \textbf{if } \#(\nstr \sym) > 0 \\
            0 & \textbf{otherwise}
        \end{cases}
    \end{equation}
    and $\typecount(\bullet, \sym)$, $\typecount(\nstr, \bullet)$ and $\typecount(\bullet, \bullet)$ are type count functions with bulleted arguments summed out. 

    \subsection{Witten--Bell Smoothing (WB) \texorpdfstring{(\citeyear{witten_bell})}{}}
    Though interpolating higher-order \ngram models with lower-order \ngram models helps handle unseen \ngram{}s, it can sometimes produce undesirable results, especially if an \ngram occurs only in a specific context. 
    A common illustrative example in the literature ~\citep{chen_empirical} is the bigram \textit{San Francisco}. 
    If the bigram \textit{San Francisco} shows up frequently in the data set, the word \textit{Francisco} will have a high unigram count, thus assigning a higher probability of continuation to \textit{Francisco} when interpolating higher-order \ngram{}s with lower order count functions. 
    This, however, may be undesirable, as the term \textit{Francisco} will rarely follow any context other than the word \textit{San}. 
    Witten--Bell smoothing addresses this by considering the number of unique continuations a context has observed, based on the intuition that backing off to lower-order  \ngram statistics is more accurate only when there are more distinct continuations of a given context. 
    The interpolation works as follows:
    \begin{equation}
        \begin{aligned}
            \pWB(\sym\mid\nstr) & = \lambda_n \qngram(\sym\mid\nstr)
            + (1-\lambda_n) \pWBminus(\sym\mid\nmstr)
        \end{aligned}
    \end{equation}
    for $\sym \in \eosalphabet$ and $\nstr \in \alphabet^{\ngr - 1}$ where
    \begin{equation}
        \begin{aligned}
            1 - \lambda_n = \frac{\typecount(\nstr, \bullet)}{\typecount(\nstr, \bullet) + \CountFun{\nstr}}.
        \end{aligned}
    \end{equation}
    Substituting, we rewrite the interpolation as
    \begin{equation}
        \begin{aligned}
            \pWB(\sym\mid\nstr) & = \frac{\CountFun{\nstr\sym} + \typecount(\nstr, \bullet)\pWBminus}{\typecount(\nstr, \bullet) +\CountFun{\nstr}}.
        \end{aligned}
    \end{equation}

    \subsection{A Log-Linear Model}
    As a simple parameter-sharing representation-based baseline, we consider a log-linear model that represents the history $\nstr$ as a concatenation of the one-hot encodings of the symbols in $\nstr$.\footnote{This exactly matches the representations used to \emph{generate} sparse representation-based \ngram LMs described in \cref{sec:n-gram-lm-generation} and thus provides a strong inductive bias for those models, making the log-linear model a strong baseline.}
    Concretely, the model represents the history $\nstr = \sym_1 \cdots \sym_{\ngr - 1}$ as
    \begin{equation}
        \encfunc{\sym_1 \cdots \sym_{\ngr - 1}} \defeq \begin{pmatrix}
            \onehot{\sym_1} \\
            \vdots \\
            \onehot{\sym_{\ngr - 1}}
        \end{pmatrix} \in \set{0, 1}^{\left(\ngr - 1\right) |\bosalphabet|}.
    \end{equation}
    Its parameter is an output matrix $\outMtx \in \R^{\nsymbols \times \left(\ngr - 1\right) |\bosalphabet|}$, which determines the logits of the conditional distribution $\pLMn\left(\sym \mid \nstr\right)$ as
    \begin{equation}
        \pLMn\left(\sym \mid \nstr\right) \defeq \softmaxfunc{\outMtx \; \encfunc{\nstr}}{\sym}.
    \end{equation}
    The model is trained by minimizing the cross-entropy loss between the true conditional distributions $\pLMn$ and the predicted distributions $\pLMn$.

    \subsection{A Neural \ngram LM}
    The neural \ngram LM is a classic neural LM popularized by \citet{NIPS2000_728f206c,10.5555/944919.944966,Bengio2006} and modernized by \citet{sun-iyyer-2021-revisiting}.
    It learns $\hiddDim'$-dimensional word2vec-style static representations of the symbols and concatenates them before feeding them through an MLP $\mlp$.\footnote{The neural model studied by \citet{sun-iyyer-2021-revisiting} adds a pooling operation over the entire preceding string. However, to keep the model truly \ngram-based, we do not include this pooling operation and ignore the symbols further than $\ngr - 1$ steps in the past.}
    The MLP thus produces the final representation of the history, which is used to define the next-symbol probabilities together with the learned output matrix $\outMtx$ as
    \begin{subequations}
        \begin{align}
            \pLM\left(\sym \mid \nstr\right) 
            & \defeq \softmaxfunc{\outMtx \; \enc\left(\strlt\right)}{\sym} \\
            & = \softmaxfunc{\outMtx \; \mlp\left(\begin{pmatrix}
                \onehot{\sym_1} \\
                \vdots \\
                \onehot{\sym_{\ngr - 1}}
            \end{pmatrix}\right)}{\sym}.
        \end{align}
    \end{subequations}
    The size of the symbol representations as well as the complexity of the MLP $\mlp$ (its depth and width) are hyperparameters of the model.

    \section{Experimental Details}

    \subsection{Data Generation} \label{sec:lm-sampling}
    \subsubsection{Generating \ngram LMs} \label{sec:n-gram-lm-generation}
    We generate the training and test datasets by randomly generating \ngram LMs and sampling strings from them.
    We construct three types of \ngram LMs:
    \begin{enumerate*}
        \item General \ngram LMs whose conditional distributions $\pLMn\left(\sym \mid \nstr\right)$ are sampled independently of each other.
        \item Sparse representation-based \ngram LMs.
        \item Dense representation-based \ngram LMs.
    \end{enumerate*}

    \paragraph{General \ngram LMs.}
    We sample general \ngram LMs by sampling, for each possible context $\nstr$ (including the contexts padded with different numbers of $\bos$ symbols), a conditional distribution $\pLNSM\left(\sym \mid \nstr\right)$ for $\sym \in \alphabet$.
    The conditional distributions are sampled from a Dirichlet distribution with concentration parameter $\alpha$, where we set $\alpha = 0.1$ to encourage concentrated distributions.
    We control the expected length of the string $\E\left[|\str|\right]$ generated by the \ngram LM by hard-coding the probability $\pLMn\left(\eos \mid \nstr\right)$ to be $\frac{1}{\E\left[|\str|\right]}$ for all $\nstr$, where we set $\E\left[|\str|\right] = 40$.
    Due to the requirement of storing all the $\nsymbols^{\ngr - 1}$ conditional distributions, we limit ourselves to the case of $\ngr \in \set{2, 4, 6}$ and $\nsymbols \in \set{8, 12, 16}$ for the general \ngram LM case.
    This procedure is described with pseudocode in \cref{alg:n-gram-lm-generation}.
    \begin{algorithm}
        \begin{algorithmic}[1]
            \Func{\textsc{Generate General \ngram LM}($\ngr, \alphabet, \alpha, \E\left[|\str|\right]$)}
            \For{$\nstr \in \bigcup_{\idxj = 0}^{\ngr - 1} \set{\bos}^\idxj \times {\alphabet^{\ngr - 1 - \idxj}}$}
            \LineComment{Iterate through all possible contexts, including $\bos$-padded contexts.}
            \State $\pLMn\left(\sym \mid \nstr\right) \sim \textsc{Dirichlet}(\alpha \one_{\nsymbols})$ for $\sym \in \alphabet$
            \State $\pLMn\left(\eos \mid \nstr\right) \gets \frac{1}{\E\left[|\str|\right]}$
            \State \textbf{Renormalize} $\pLMn\left(\eossym \mid \nstr\right)$ for $\eossym \in \eosalphabet$
            \EndFor
            \State \Return $\pLMn$
            \EndFunc
        \end{algorithmic}
        \caption{The generation of a random general \ngram LM.}
        \label{alg:n-gram-lm-generation}
    \end{algorithm}

    \paragraph{Representation-based \ngram LMs.}
    Representation-based \ngram LMs are generated by defining the conditional distributions $\pLMn\left(\sym \mid \nstr\right)$ in terms of an \defn{output matrix} $\outMtx$ which transforms the vectorial \defn{representations} $\encfunc{\nstr} \in \R^{\hiddDim}$ of the history $\nstr$ into the (logits of the) conditional distribution $\pLMn\left(\sym \mid \nstr\right)$.
    More concretely, let $\enc \colon \alphabet^{\ngr - 1} \to \R^{\hiddDim}$ be an representation function that maps the history $\nstr$ to a vector in $\R^{\hiddDim}$. 
    Then, we define the conditional distribution $\pLMn\left(\sym \mid \nstr\right)$ for $\sym \in \alphabet$ as
    \begin{equation}
        \pLMn\left(\sym \mid \nstr\right) \defeq \softmaxfunc{\outMtx \; \encfunc{\nstr}}{\sym},
    \end{equation}
    where $\outMtx \in \R^{\nsymbols \times \hiddDim}$ is an output matrix.
    Once again, we hard-code the probability of the end-of-string symbol $\eos$ to be $\frac{1}{\E\left[|\str|\right]} = \frac{1}{40}$ for all $\nstr$ by post-hoc renormalizing the conditional distributions at every time step.

    We consider two representation functions $\enc$:
    \begin{enumerate}
        \item \defn{Sparse}: We define the sparse representation function $\enc\colon \alphabet^{\ngr - 1} \to \set{0, 1}^{\left(\ngr - 1\right) |\bosalphabet|}$ as the function mapping the history $\sym_1 \cdots \sym_{\ngr - 1}$ to the concatenation of the one-hot encodings of its symbols.
        In symbols, this mean
        \begin{equation}
            \encfunc{\sym_1 \cdots \sym_{\ngr - 1}} \defeq \begin{pmatrix}
                \onehot{\sym_1} \\
                \vdots \\
                \onehot{\sym_{\ngr - 1}}
            \end{pmatrix} \in \set{0, 1}^{\left(\ngr - 1\right) |\bosalphabet|}
        \end{equation}
        where $\onehot{\sym} \in \set{0, 1}^{|\bosalphabet|}$ is the one-hot encoding of the symbol $\sym$.
        Notice that in this case, the size of the representations grows with the size of the alphabet.
        \item \defn{Dense}: We define a dense representation function $\enc\colon \alphabet^{\ngr - 1} \to \R^{\left(\ngr - 1\right) \hiddDim'}$ as the function mapping the history $\sym_1 \cdots \sym_{\ngr - 1}$ to the concatenation of \emph{dense} embeddings $\inEmbeddingFun{\sym} \in \R^{\hiddDim'}$ of its symbols.
        We generate the symbol embeddings randomly from a standard normal distribution.
        We use $\hiddDim' = 16$ for the dense representation of symbols.
        Furthermore, since the \emph{rank} of the output matrix $\outMtx$ was found to be a significant predictor of the learnability of general regular LMs \citep{borenstein2024languages}, we control for the rank of the output matrix generating $\outMtx$ by constructing $\outMtx = \outMtx_1 \outMtx_2$ the product of two random matrices of $\outMtx_1  \in \R^{\nsymbols \times \rank}$ and $\outMtx_2 \in \R^{\rank \times \hiddDim}$, resulting in a rank-$\rank$ matrix of size $\nsymbols \times \hiddDim$, where we vary the rank as $\rank \in \set{2, 8, 16}$.
        This results in a fixed-size representation of the history, regardless of the size of the alphabet.
    \end{enumerate}
    The randomly-generated matrices ($\outMtx$ in the sparse case and $\outMtx_1$ and $\outMtx_2$ in the dense case) are generated by sampling their entries independently from a standard normal distribution.

    The main difference between sparse and dense representation functions is therefore the degree of parameter sharing between the different symbols in the history.
    Since, unlike in the general \ngram LM case, we do not need to store the conditional distributions for all possible histories, we can consider larger values of $\ngr$ and $\nsymbols$.
    In particular, we generate datasets with $\ngr \in \set{4, 8, 12}$ and $\nsymbols \in \set{64, 128, 256}$.

    \subsubsection{Generating Train and Test Datasets}
    The training and test datasets are generated by sampling strings from the \ngram LMs.
    Let $\pLMn$ be an \ngram LM.
    We sample \emph{disjoint} training and test datasets from $\pLMn$ in a multi-step process:
    \begin{enumerate}[label=(\arabic*.)]
        \item Sample a large set of strings $\dataset'$ from $\pLMn$ (where strings are not repeated).
        \item Divide $\dataset' = \dataset'_\text{Train} \sqcup \dataset'_\text{Train}$ into the set of strings $\dataset'_\text{Train}$ that are allowed to appear in the training dataset and a set of strings $\dataset'_\text{Train}$ that are allowed to appear in the test dataset.
        \item Sample two large (multi-)sets of strings $\datasetTrain$ and $\datasetTest$ from $\pLMn$.
        \item Remove all strings from $\datasetTrain$ that are in $\dataset'_\text{Test}$ and all strings from $\datasetTest$ that are in $\dataset'_\text{Train}$.
        \item Retain $\datasetSizeTrain$ strings from $\datasetTrain$ and $\datasetSizeTest$ strings from $\datasetTest$.
    \end{enumerate}
    This procedure ensures that the training and test datasets are disjoint and that the test dataset is not seen during training.

\begin{figure}
        \centering
        \begin{subfigure}{\textwidth}
            \centering
            \includegraphics[width=\textwidth]{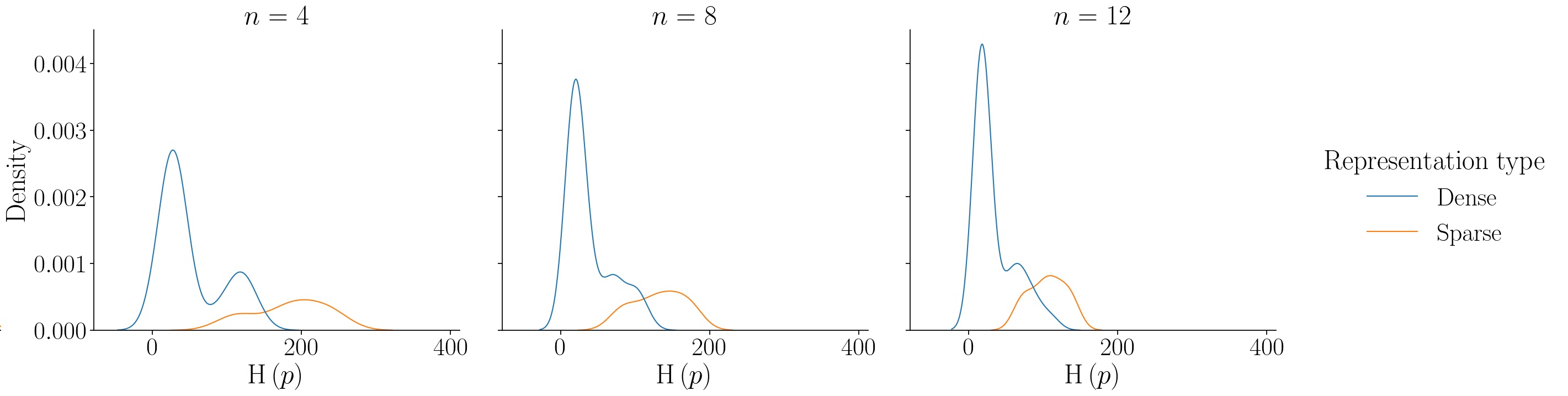}
            \caption{Distribution of the entropy of the datasets from representation-based \ngram LMs.}
        \end{subfigure}
        \begin{subfigure}{\textwidth}
            \centering
            \includegraphics[width=0.7\textwidth]{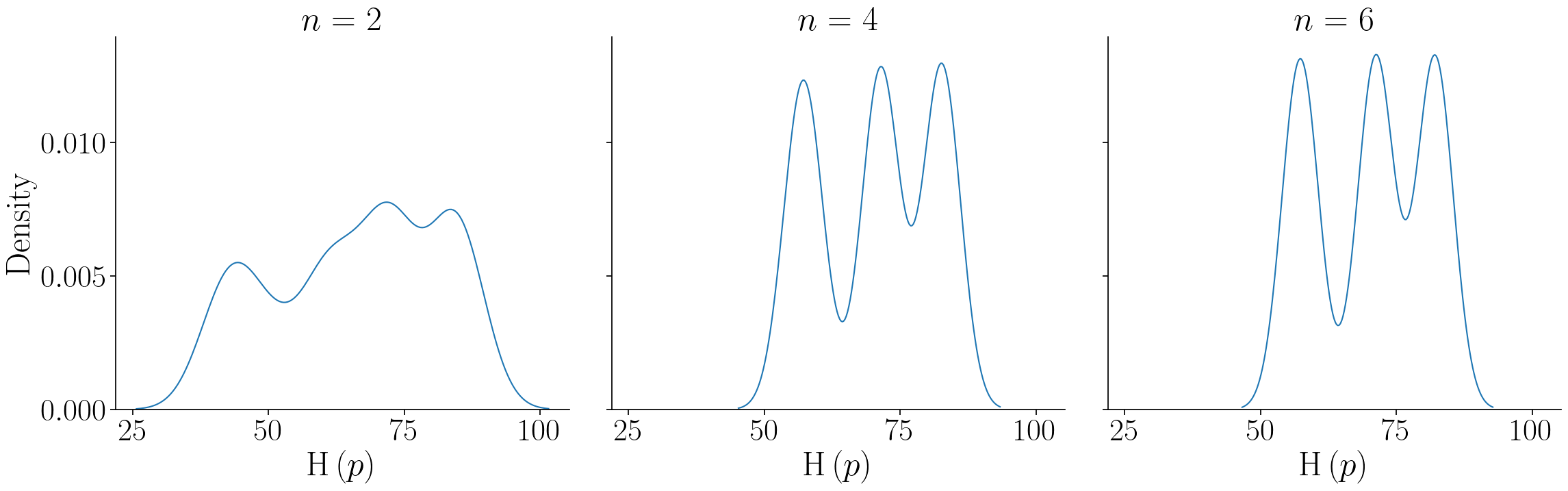}
            \caption{Distribution of the entropy of the datasets from the general \ngram LMs.}
        \end{subfigure}
        \caption{Distributions of the mean string lengths and the LM entropies for the $45$ generated datasets.}
        \label{fig:dataset-stats}
    \end{figure}

    \cref{fig:dataset-stats} shows the distributions of the entropies for the generated datasets.

    \subsection{Models} \label{sec:learning}

    \subsubsection{Transformer Models} \label{sec:implementation-transformer-models} 
    We use the GPT-2 model architecture \cite{radford2019language} of varying sizes. We sample the number of heads $\tfheadnum$ the number of layers $\tfnumlayer$ from $\set{1, \ngr}$, where $\ngr$ is the learnt \ngram{}'s order. We also control for the type of attention activation function, using either the standard $\softmax$ or its variant $\sparsemax$.  Moreover, we use an embedding size of $256$, a hidden representation size of $512$, and an output size (the dimensionality of the final linear layer) of $128$. The number of trainable parameters of the models ranges between $500$k and $4$M. We train each model for $10$ epochs with an input context length of $256$ using a batch size of $128$ and a learning rate of $0.0001$, an AdamW \citep{loshchilov2018fixing} optimizer with default settings, and a standard cross-entropy loss. We did not tie the weights of the word embeddings. Training a single instance took, on average, $30$ minutes, with some variations depending on the size of the dataset and exact architecture size. 

    \subsubsection{Smoothing Techniques} \label{sec:implementation-classical} 
    We train a number of smoothing techniques on the generated datasets as a baseline for the transformers.
    We use the smoothing techniques described in \cref{sec:estimation-techniques}.
    Each of them allows us to control the order of the learned \ngram LM, $\nhat$.
    For each setting, we test out three scenarios: An under-parametrized model ($\nhat = \ngr - 2$), a well-parametrized model ($\nhat = \ngr$), and an over-parametrized model ($\nhat = 2 \ngr$ for $\ngr \in \set{4, 8}$ and $\nhat = 20$ for $\ngr = 12$).\footnote{While transformer models do not have an analogous hyperparameter to make them fit the ground-truth \ngram LM perfectly, we vary their sizes (number of heads and layers) to find the best-performing model, as motivated by theoretical work.}
    This results in models between with a large number of parameters---the largest model on $\nsymbols = 256$ symbols of order $\nhat = 20$ defines in the order of $10^48$ parameters---but note that only the parameters corresponding to observed \ngram{}s are memorized, making the training feasible.
    % Intuitively, smoothing techniques are especially useful in the over-parametrized setting, where they can help to generalize the model to unseen $\nhat$-grams.\anej{remove}
    Some of the estimation techniques---MLE and Witten--Bell---do not have any hyperparameters and are trained with the default settings.
    For the other techniques (Add-$\lambda$ smoothing and Absolute discounting), we train models with the hyperparameters listed in \cref{tab:smoothing-hyperparameters} and report the performance of the best-performing models.
    \begin{table}
        \centering
        \begin{tabular}{lcc}
            \toprule
            \textbf{Smoothing Technique} & \textbf{Hyperparameter} & \textbf{Values}  \\
            \midrule
            Add-$\lambda$ smoothing (Add-$\lambda$)           & $\lambda$               & $0.01, 0.1, 1$            \\
            Absolute Discounting (AD)        & $\delta$                & $0.6, 0.8, 0.95$ \\
            \bottomrule
        \end{tabular}
        \caption{Hyperparameters for the smoothing techniques.}
        \label{tab:smoothing-hyperparameters}
    \end{table}
    We use the standard \texttt{nltk} implementations \citep{bird2009natural} for these estimation techniques.

    \subsubsection{Log-Linear Model and Neural \ngram LM} \label{sec:implementation-log-linear}

    We implemented the log-linear model and the neural \ngram LM in PyTorch \citep{paszke2019pytorch}.
    The log-linear model only learns the output matrix $\outMtx$ while our implementation of the neural LM closely follows the model defined by \citet{NIPS2000_728f206c,10.5555/944919.944966,Bengio2006} with the modern improvements studied by \citet{sun-iyyer-2021-revisiting}.\footnote{Since we are interested in \ngram LMs, we do not implement the pooling of the longer context from \citet{sun-iyyer-2021-revisiting}.}
    As with the smoothing techniques, we study the effects of the order $\nhat$ of the trained LM on the performance of the log-linear model and the neural \ngram LM.
    Thus, we fit models with $\nhat \in \set{\ngr - 2, \ngr, \min(2 \ngr, 20)}$, resulting with models with between $4$k and $1.25$M parameters.
    The models are trained by minimizing the cross-entropy loss between the empirical distribution of the training dataset and the predicted distributions $\pLM$.
    For the log-linear model, we use the Adam optimizer \citep{kingma2017adam} with a learning rate of $0.1$ and a batch size of $1024$.
    We train the models for $16$ epochs and halve the learning rate after every five epochs.
    The neural \ngram LMs were implemented as a shallow neural network with three learned layers. The first is an embedding later with a dimensionality of $128$. The second is a fully connected layer with an output dimensionality of $512$, ReLU activation function, and dropout probability of $0.5$. The third layer is a softmax-normalized linear transformation for predicting next-symbol probabilities, i.e., the matrix $\outMtx$. The number of trainable parameters of the models ranges between $400$k and $1.5$M. They were trained for 20 epochs with early stopping on a development set that was sampled from the training set ($80\%$--$20\%$ split) using a batch size of $128$ and, learning rate of $5e^{-5}$, and Adam optimizer with default parameters. Training of a single instance took, on average, 5 minutes, with some variations depending on the size of the dataset and model. 

    \subsection{Evaluation} \label{sec:evaluation}
    As described in \cref{sec:framework}, we evaluate the learned LMs by computing the KL divergence between the learned LM $\pLM$ and the ground-truth \ngram LM $\pLMn$.
    Concretely, we can compute $\KLFun{\pLMn}{\pLM}$ through its decomposition as
    \begin{equation} 
        \KLFun{\pLMn}{\pLM} = \entropy(\pLMn, \pLM) - \entropy(\pLMn). \label{eq:entropy-diff}
    \end{equation}
    While $\entropy\left(\pLMn\right)$ can be computed analytically with a dynamic program \citep{eisner-2002-parameter,zmigrod-etal-2021-efficient-computation}, the runtime complexity of $\bigO{\left(\eosnsymbols^{\ngr - 1}\right)^3}$ makes exact computations infeasible for even moderately large $\eosnsymbols$ and $\ngr$.
    We thus rely on empirical estimates of both terms in \cref{eq:entropy-diff} on the test dataset.
    Concretely, we compute
    \begin{subequations}
        \begin{align}
            \widehat{\entropy}\left(\pLMn\right)         & = -\frac{1}{\datasetSizeTest} \sum_{\str \in \datasetTest} \log \pLMn\left(\str\right)   \\
            \widehat{\entropy}\left(\pLMn, \pLM\right) & = -\frac{1}{\datasetSizeTest} \sum_{\str \in \datasetTest} \log \pLM\left(\str\right),
        \end{align}
    \end{subequations}
    which allows us to compute the empirical approximation of $\KLFun{\pLMn}{\pLM}$, 
    \begin{equation}
        \widehat{\KL}\left(\pLMn \mid \mid \pLM\right) \defeq \widehat{\entropy}(\pLMn, \pLM) - \widehat{\entropy}(\pLMn).
    \end{equation}

    \subsection{Statistical Analysis} \label{sec:stat-analysis} 
    To assess the influence of the predictors specified in \Cref{tab:predictors} on the learnability of the \ngram LMs (i.e, the empirical KL divergence on the test dataset), we implement a linear regression model predicting the KL divergence based on the predictors. 
    Before fitting the model, we standardize each parameter with a $z$-score transformation for an interpretable comparison of the estimated coefficients.

    The linear regression model provides insights into the influence of each predictor on the dependent variable (KL divergence) by assigning each parameter three key values: the coefficient of the linear model $\hat{\beta}_i$, the standard deviation of the coefficient, and a $p$-value. 
    The magnitude of the coefficient $\hat{\beta}_i$ indicates the strength of the predictor's effect on the dependent variable.
    The coefficient's \emph{sign} reveals whether this effect is positive (an increase in the predictor is expected to increase the value of the parameter; in our case, this indicates that the increase of the parameter is associated with a \emph{worse} performance of the model and vice-versa) or negative (an increase in the predictor is expected to decrease the value of the parameter; in our case, this indicates that the increase of the parameter is associated with a \emph{better} performance of the model and vice-versa). 
    The standard deviation measures the variability or uncertainty of the $\hat{\beta}_i$ coefficient, providing a sense of the reliability of this estimate. 
    The $p$-value quantifies the statistical significance of the effect, indicating the likelihood that the observed relationship occurred by chance.
    It thus provides a measure of the reliability of the effect, with lower $p$-values indicating a more reliable effect.

    \section{Additional Results} \label{sec:additional-results}

    \subsection{The Effect of Parameter Sharing} \label{sec:additional-results-onehot-vs-vectorial}

    In \cref{sec:results-parameter-sharing} (particularly, in \cref{tab:general-vs-representations-results}), we compared the performance of the models of interest across general and dense representation-based \ngram LMs. 
    As expected, neural LMs fare much better with representation-based \ngram LMs, whereas the difference in the performance of counting-based methods is smaller.
    \cref{tab:appendix-general-vs-representations-general-results,tab:appendix-general-vs-representations-representations-results} provide results for additional model orders and alphabet sizes, showing the same trends as \cref{tab:general-vs-representations-results}---all learning methods perform better on representation-based \ngram LMs across all settings.
    
    \begin{table*}[!ht]
        \centering
        \begin{subtable}{\textwidth}            
        \centering
        \begin{tabular}{cccccccccc}
                \toprule
                        $\ngr$            & \multicolumn{3}{c}{$2$}         & \multicolumn{3}{c}{$4$}          & \multicolumn{3}{c}{$6$}                                                                                                                                                                                                                   \\
                \cmidrule(lr){2-4} \cmidrule(lr){5-7} \cmidrule(lr){8-10}
                                $\nsymbols$      & $8$                             & $12$                             & $16$
                                    & $8$                             & $12$                             & $16$
                                    & $8$                             & $12$                             & $16$                                                                                                                                                                                                                                      \\
                \midrule
            \textbf{Classic}           & \meanStd{-0.08}{0.94} & \meanStd{-0.35}{0.78} & \meanStd{0.18}{1.49} & \meanStd{-1.00}{0.39} & \meanStd{0.23}{1.41} & \meanStd{1.86}{1.96} & \meanStd{3.04}{1.46} & \meanStd{17.34}{1.01} & \meanStd{50.35}{1.87}         \\
            \textbf{LL} & \meanStd{35.31}{7.55} & \meanStd{50.57}{12.27} & \meanStd{59.76}{4.96} & \meanStd{86.03}{2.05} & \meanStd{103.36}{2.95} & \meanStd{111.95}{5.41} & \meanStd{101.42}{3.46} & \meanStd{109.60}{2.06} & \meanStd{117.83}{2.37}
            \\
            \textbf{Neural}            & \meanStd{-0.19}{1.04} & \meanStd{-0.46}{0.81} & \meanStd{0.07}{1.57} & \meanStd{0.91}{0.51} & \meanStd{16.29}{1.74} & \meanStd{40.00}{2.78} & \meanStd{60.87}{2.87} & \meanStd{79.77}{1.63} & \meanStd{90.01}{1.18}  \\
            \midrule
            \textbf{TF}             & \meanStd{0.07}{1.09} & \meanStd{-0.48}{0.82} & \meanStd{0.18}{1.50} & \meanStd{0.18}{0.53} & \meanStd{4.68}{1.61} & \meanStd{10.98}{2.35} & \meanStd{67.06}{2.91} & \meanStd{77.95}{1.74} & \meanStd{86.38}{1.88}         \\
            \bottomrule
        \end{tabular}
        \caption{Learnability of general \ngram LMs.}
        \label{tab:appendix-general-vs-representations-general-results}
        \end{subtable}
        \begin{subtable}{\textwidth}
            \centering
            \begin{tabular}{cccccccccc}
                \toprule
                        $\ngr$            & \multicolumn{3}{c}{$2$}         & \multicolumn{3}{c}{$4$}          & \multicolumn{3}{c}{$6$}                                                                                                                                                                                                                   \\
                \cmidrule(lr){2-4} \cmidrule(lr){5-7} \cmidrule(lr){8-10}
                                $\nsymbols$      & $8$                             & $12$                             & $16$
                                    & $8$                             & $12$                             & $16$
                                    & $8$                             & $12$                             & $16$                                                                                                                                                                                                                                      \\
                \midrule
                \textbf{Classic}           & \meanStd{-0.25}{0.53} & \meanStd{-0.35}{1.47} & \meanStd{0.55}{1.00} & \meanStd{0.72}{0.66} & \meanStd{0.89}{0.69} & \meanStd{1.28}{0.90} & \meanStd{2.36}{0.23} & \meanStd{2.45}{0.96} & \meanStd{3.11}{0.42}         \\
                \textbf{LL} & \meanStd{25.23}{9.12} & \meanStd{26.83}{9.12} & \meanStd{29.10}{7.82} & \meanStd{19.67}{4.28} & \meanStd{24.03}{5.47} & \meanStd{26.55}{6.62} & \meanStd{21.96}{3.38} & \meanStd{27.10}{4.95} & \meanStd{28.37}{3.53}
                \\
                \textbf{Neural}            & \meanStd{-0.39}{0.60} & \meanStd{-0.52}{1.73} & \meanStd{0.26}{0.41} & \meanStd{0.61}{0.65} & \meanStd{0.35}{0.66} & \meanStd{1.15}{0.72} & \meanStd{1.50}{0.20} & \meanStd{1.43}{0.78} & \meanStd{1.63}{0.47}  \\
                \midrule
                \textbf{TF}             & \meanStd{1.45}{2.34} & \meanStd{1.58}{3.83} & \meanStd{1.34}{0.35} & \meanStd{4.55}{3.02} & \meanStd{1.82}{0.49} & \meanStd{3.46}{1.15} & \meanStd{4.63}{2.00} & \meanStd{2.80}{0.72} & \meanStd{3.68}{1.33}  \\
                \bottomrule
            \end{tabular}
            \caption{Learnability of small representation-based \ngram LMs.}
            \label{tab:appendix-general-vs-representations-representations-results}
        \end{subtable}
        \caption{Learnability of general and (small) representation based \ngram LMs.}
    \end{table*}

    % \begin{table*}[!ht]
    %     \centering
    %     \begin{tabular}{ccccccc}
    %         \toprule
    %         $\nsymbols$ & \multicolumn{2}{c}{$64$} & \multicolumn{2}{c}{$128$} & \multicolumn{2}{c}{$256$} \\
    %         \cmidrule(lr){2-3} \cmidrule(lr){4-5} \cmidrule(lr){6-7} Dense & No & Yes & No & Yes & No & Yes \\
    %         \midrule
    %         \textbf{Classic}           & \meanStd{104.57}{5.87} & \meanStd{49.41}{8.95} & \meanStd{130.03}{6.96} & \meanStd{68.35}{5.64} & \meanStd{141.89}{6.28} & \meanStd{86.94}{6.74}         \\
    %         \textbf{LL} & \meanStd{32.43}{3.13} & \meanStd{43.40}{5.46} & \meanStd{40.01}{3.97} & \meanStd{49.75}{4.12} & \meanStd{51.14}{4.73} & \meanStd{61.60}{5.45}
    %         \\
    %         \textbf{Neural}            & \meanStd{}{} & \meanStd{}{} & \meanStd{}{} & \meanStd{}{} & \meanStd{}{} & \meanStd{}{}  \\
    %         \midrule
    %         \textbf{TF}             & \meanStd{12.03}{4.03} & \meanStd{13.96}{1.93} & \meanStd{58.09}{7.38} & \meanStd{19.01}{2.64} & \meanStd{116.07}{5.43} & \meanStd{25.15}{2.33}  \\
    %     \bottomrule
    %     \end{tabular}
    %     \caption{Learnability of sparse and dense representation-based \ngram LMs for $\ngr = 12$.}
    %     \label{tab:onehot-vs-vectorial-results}
    % \end{table*}

    The next natural question is whether the degree to which the parameters are shared affects the performance as well.
    To determine that, \cref{tab:onehot-vs-vectorial-onehot-results} shows the performance on one-hot representation-based \ngram LMs.
    For an easier comparison to the results on dense representation-based \ngram LMs, the results from \cref{tab:best-results} are reproduced in \cref{tab:onehot-vs-vectorial-vectorial-results}.
    Transformers perform better on dense representation-based \ngram LMs across all settings, as expected, since those more closely fit their modeling assumptions.
    In fact, transformers do not perform much better than classical smoothing techniques on some of the configurations.
    While dense-representation-based models are also better learned by smoothing methods, the log-linear model interestingly performs \emph{better} on the sparse \ngram LMs (apart from $\ngr = 4$).
    This is in line with the specification of the log linear model: The model \emph{a priori} assumes a sparse-representation-based \ngram LM, and thus fits the model specifications well, which again shows the effect of correct model specification.
    
    \begin{table*}[!ht]
        \centering
        \begin{subtable}{\textwidth}
        \centering
        \begin{tabular}{cccccccccc}
            \toprule
                    $\ngr$            & \multicolumn{3}{c}{$4$}          & \multicolumn{3}{c}{$8$} & \multicolumn{3}{c}{$12$}                                                                                                                                                                                                                   \\
                    \cmidrule{2-4} \cmidrule{5-7} \cmidrule{8-10} 
                            $\nsymbols$ & $64$ & $128$ & $256$ & $64$ & $128$ & $256$ & $64$ & $128$ & $256$ \\
            \midrule
            \textbf{Classic}           & \meanStd{16.78}{3.99} & \meanStd{28.67}{2.20} & \meanStd{47.18}{7.87} & \meanStd{76.08}{8.97} & \meanStd{94.45}{7.19} & \meanStd{112.85}{8.70} & \meanStd{104.57}{5.87} & \meanStd{130.03}{6.96} & \meanStd{141.89}{6.28}         \\
            \textbf{LL} & \meanStd{40.36}{5.40} & \meanStd{42.44}{2.34} & \meanStd{56.95}{8.99} & \meanStd{36.33}{8.23} & \meanStd{40.95}{5.66} & \meanStd{53.01}{7.57} & \meanStd{32.43}{3.13} & \meanStd{40.01}{3.97} & \meanStd{51.14}{4.73}
            \\
            \textbf{Neural}            & \meanStd{\mathbf{-1.14}}{3.81} & \meanStd{\mathbf{0.84}}{1.91} & \meanStd{\mathbf{17.82}}{7.08} & \meanStd{\mathbf{5.27}}{6.44} & \meanStd{\mathbf{13.08}}{5.27} & \meanStd{\mathbf{35.84}}{6.68} & \meanStd{\mathbf{7.61}}{2.91} & \meanStd{\mathbf{22.48}}{3.76} & \meanStd{\mathbf{45.11}}{4.15}  \\
            \midrule
            \textbf{TF}          & \meanStd{0.63}{3.73} & \meanStd{6.90}{1.93} & \meanStd{35.03}{7.47} & \meanStd{10.36}{6.95} & \meanStd{40.33}{9.09} & \meanStd{98.79}{10.55} & \meanStd{14.38}{3.77} & \meanStd{70.50}{14.64} & \meanStd{118.36}{5.80} \\
            \bottomrule
        \end{tabular}
        \caption{Learnability of sparse representation-based \ngram LMs with $\rank = 16$.}
        \label{tab:onehot-vs-vectorial-onehot-results}
        \end{subtable}
        \begin{subtable}{\textwidth}
        \centering
        \begin{tabular}{lccccccccc}
            \toprule
            $\ngr$           & \multicolumn{3}{c}{$4$}          & \multicolumn{3}{c}{$8$} & \multicolumn{3}{c}{$12$}                                                                                                                                                                                                                   \\
            \cmidrule(lr){2-4} \cmidrule(lr){5-7} \cmidrule(lr){8-10}
            $\nsymbols$ & $64$ & $128$ & $256$ & $64$ & $128$ & $256$ & $64$ & $128$ & $256$ \\
            \midrule
            % \textbf{MLE}           & \meanStd{3.25}{0.58} & \meanStd{6.57}{1.29} & \meanStd{12.05}{1.86} & \meanStd{73.70}{41.01} & \meanStd{235.19}{32.65} & \meanStd{298.84}{30.01} & \meanStd{245.39}{57.65} & \meanStd{300.87}{26.61} & \meanStd{340.88}{33.70}         \\
            \textbf{Classic}            & \meanStd{2.11}{0.39} & \meanStd{3.40}{0.81} & \meanStd{5.00}{0.72} & \meanStd{25.72}{8.48} & \meanStd{54.95}{4.17} & \meanStd{67.64}{5.25} & \meanStd{58.17}{9.36} & \meanStd{74.09}{6.78} & \meanStd{89.90}{6.96}  \\
            \textbf{LL} & \meanStd{32.05}{3.14} & \meanStd{37.10}{9.53} & \meanStd{40.61}{3.93} & \meanStd{40.45}{11.01} & \meanStd{58.26}{3.64} & \meanStd{62.00}{5.23} & \meanStd{43.40}{5.46} & \meanStd{49.75}{4.12} & \meanStd{61.60}{5.45}
            \\
            \textbf{Neural}            & \meanStd{\mathbf{1.14}}{0.64} & \meanStd{\mathbf{1.98}}{0.91} & \meanStd{\mathbf{2.17}}{1.46} & \meanStd{\mathbf{5.96}}{1.86} & \meanStd{\mathbf{9.13}}{0.88} & \meanStd{\mathbf{9.06}}{0.68} & \meanStd{\mathbf{7.71}}{0.75} & \meanStd{\mathbf{9.87}}{1.28} & \meanStd{\mathbf{11.20}}{1.09}  \\
            \midrule
            \textbf{TF}             & \meanStd{2.43}{0.51} & \meanStd{5.04}{1.75} & \meanStd{5.63}{1.80} & \meanStd{9.52}{1.98} & \meanStd{14.72}{0.95} & \meanStd{16.89}{1.33} & \meanStd{14.73}{1.70} & \meanStd{22.79}{6.42} & \meanStd{33.99}{9.06}  \\
            \bottomrule
        \end{tabular}
        \caption{Learnability of dense representation-based \ngram LMs with $\rank = 16$.}
        \label{tab:onehot-vs-vectorial-vectorial-results}
        \end{subtable}
        \caption{Learnability of sparse and dense representation-based \ngram LMs.}
    \end{table*}	

    \subsection{The Effect of the Rank} \label{sec:rank-results}
    In \cref{sec:results-complexity}, we investigated the trends in the performance with respect to the order of the \ngram model and the size of the alphabet (cf. \cref{tab:best-results}). 
    Here, we also consider the rank $\rank$ of $\outMtx$.
    The results for varying ranks are presented in \cref{tab:rank-results}.
    They again follow the intuitions from theory.
    The performance smoothing methods and the log-linear model, which can by design implement any-rank dense-representation-based \ngram LM, is unaffected by $\rank$.
    The performance of neural \ngram{}s and transformers (which in our experiments all have rank at most $128$---the output dimension), however, is negatively correlated with the rank. 
    This is confirmed by the analysis of the predictors of the transformer performance (cf. \cref{tab:tf-coefficients}), although the effect is not as significant and strong as for the other predictors.

    \begin{table*}[!ht]
        \centering
        \begin{tabular}{cccccccccc}
            \toprule
            $\nsymbols$ & \multicolumn{3}{c}{$64$} & \multicolumn{3}{c}{$128$} & \multicolumn{3}{c}{$256$} \\
            \cmidrule(lr){2-4} \cmidrule(lr){5-7} \cmidrule(lr){8-10} $\rank$ & 2 & 8 & 16 & 2 & 8 & 16 & 2 & 8 & 16 \\
            \midrule
            \textbf{Classic}           & \meanStd{75.82}{3.52} & \meanStd{72.86}{5.39} & \meanStd{49.41}{8.95} & \meanStd{79.71}{9.06} & \meanStd{82.67}{4.94} & \meanStd{68.35}{5.64} & \meanStd{79.21}{6.90} & \meanStd{94.94}{8.97} & \meanStd{86.94}{6.74}         \\
            \textbf{LL} & \meanStd{45.82}{3.52} & \meanStd{47.98}{4.49} & \meanStd{43.40}{5.46} & \meanStd{48.94}{4.10} & \meanStd{54.77}{4.84} & \meanStd{49.75}{4.12} & \meanStd{61.66}{9.63} & \meanStd{60.69}{5.13} & \meanStd{61.60}{5.45} 
            \\
            \textbf{Neural}            & \meanStd{2.28}{2.48} & \meanStd{4.54}{1.02} & \meanStd{7.71}{0.75} & \meanStd{2.16}{4.03} & \meanStd{5.08}{1.04} & \meanStd{9.87}{1.28} & \meanStd{0.31}{4.79} & \meanStd{7.34}{1.55} & \meanStd{11.20}{1.09}  \\
            \midrule
            \textbf{TF}             & \meanStd{5.49}{24.78} & \meanStd{9.06}{3.57} & \meanStd{13.96}{1.93} & \meanStd{5.90}{11.60} & \meanStd{11.05}{1.61} & \meanStd{19.01}{2.64} & \meanStd{4.34}{30.98} & \meanStd{14.30}{2.70} & \meanStd{25.15}{2.33}  \\
        \bottomrule
        \end{tabular}
        \caption{The effect of the rank $\rank$ on the performance of the best performing models for $\ngr = 12$.}
        \label{tab:rank-results}
    \end{table*}

    \subsection{The Effect of Sparse Attention} \label{sec:sparse-attention-results}

    \cref{tab:tf-coefficients} suggests a significant impact of the use of sparse attention on the ability to learn \ngram LMs. 
    This is confirmed by looking at the difference in the performance of soft- and sparse-attention transformers in \cref{tab:softmax-vs-sparsemax-results}.
    
    \begin{table*}
        \centering
        \begin{tabular}{cccccccccc}
            \toprule
                    $\ngr$            & \multicolumn{3}{c}{$4$}          & \multicolumn{3}{c}{$8$} & \multicolumn{3}{c}{$12$}                                                                                                                                                                                                                   \\
            \cmidrule(lr){2-4} \cmidrule(lr){5-7} \cmidrule(lr){8-10} 
                            $\nsymbols$ & $64$ & $128$ & $256$ & $64$ & $128$ & $256$ & $64$ & $128$ & $256$ \\
            \midrule
            $\softmax$               &  \meanStd{2.58}{0.52} & \meanStd{5.45}{1.50} & \meanStd{6.52}{1.73} & \meanStd{14.47}{5.67} & \meanStd{22.37}{7.95} & \meanStd{23.51}{8.49} & \meanStd{23.74}{8.82} & \meanStd{37.52}{7.51} & \meanStd{47.79}{7.64}  \\
            $\sparsemax$               &  \meanStd{\mathbf{2.43}}{0.51} & \meanStd{\mathbf{5.04}}{1.75} & \meanStd{\mathbf{5.63}}{1.80} & \meanStd{\mathbf{9.52}}{1.98} & \meanStd{\mathbf{14.72}}{0.95} & \meanStd{\mathbf{16.89}}{1.33} & \meanStd{\mathbf{14.73}}{1.70} & \meanStd{\mathbf{22.79}}{6.42} & \meanStd{\mathbf{33.99}}{9.06}  \\
            \bottomrule
        \end{tabular}
        \caption{Performance of soft- and sparse-attention transformers on dense representation \ngram LMs with $\rank = 16$.}
        \label{tab:softmax-vs-sparsemax-results}
    \end{table*}

    \subsection{The Effect of Over-parametrization}

    The baselines used in the experiments in the main part of the paper (classic smoothing techniques, the log-linear model, and the neural \ngram model) are particularly well-specified for the LMs they approximate---in particular, they were trained with the correct order of the ground-truth \ngram LM, $\ngr$.
    Apart from the intuitions offered by the theoretical constructions, such an appropriate parametrization is more difficult to specify for transformers, whose parameters do not match the \ngram definition as closely. 
    In this section, we investigate how much possible misspecification of the baselines impacts their performance and compare it to the performance of the largest and smallest transformers trained.
    
    To test the effect of \emph{over-parametrization}, that is, assuming a too-large order $\nhat$, \cref{tab:over-parametrization} shows the performance of the best-performing baseline models with $\nhat = 2 \ngr$ for $\ngr \in \set{4, 8}$ and $\nhat = 20$ for $\ngr = 12$.
    Again, \cref{tab:onehot-vs-vectorial-vectorial-results} serves as reference for the performance of the best models.
    \cref{tab:over-parametrization} also shows the performance of the largest transformer models for each of the settings (that is, one where both the number of layers and heads are largest).\footnote{According to the theoretical constructions by \citet{svete2024transformers}, such transformers are over-parametrized---only one of the number of heads or layers needs to increase.}
    Over-parameterization is particularly harmful to models whose number of parameters increases most---the classical smoothing techniques.
    Other models are less affected.
    In particular, the neural \ngram model performs as well as the optimally-parametrized variant, and transformers remain close to their best performance, with the exception of the largest model.

    We contrast this to the effects of \emph{under-parameterization}, where we set $\nhat = \ngr - 2$ and constrain the transformers to a single head and layer.
    The results of these runs are presented in \cref{tab:under-parametrization}.
    Under-parametrization noticeably degrades the performance of most models and is most noticeable with the smaller orders $\ngr$. 
    Due to the fast growth of the number of parameters in the classical models, under-parametrization actually outperforms over-parameterized models, likely due to the parameter-sharing nature of the ground-truth \ngram LMs, possibly making them more easily approximatable with lower-order models.
    Single-head and single-layer transformers, in contrast, perform noticeably worse than their well- or over-parametrized variants, again in line with the intuitions from the theoretical constructions.
    
    \begin{table*}[!ht]
        \centering
        \begin{subtable}{\textwidth}
        \centering
        \begin{tabular}{ccccccccccccccccc}
            \toprule
                    $\ngr$            & \multicolumn{3}{c}{$4$} & \multicolumn{3}{c}{$8$} & \multicolumn{3}{c}{$12$}                                                                                                                                                                                                                   \\
            \cmidrule{2-4} \cmidrule(lr){5-7} \cmidrule(lr){8-10} 
                            $\nsymbols$ & $64$ & $128$ & $256$ & $64$ & $128$ & $256$ & $64$ & $128$ & $256$ \\
            \midrule
            \textbf{Classic}            & \meanStd{3.49}{0.30} & \meanStd{7.30}{1.55} & \meanStd{10.21}{0.75} & \meanStd{29.98}{9.82} & \meanStd{61.97}{4.51} & \meanStd{73.69}{5.46} & \meanStd{50.22}{9.03} & \meanStd{68.50}{5.64} & \meanStd{87.01}{6.85}  \\
            \textbf{LL}  & \meanStd{33.69}{3.14} & \meanStd{40.24}{9.53} & \meanStd{43.49}{4.84} & \meanStd{63.38}{14.44} & \meanStd{83.96}{9.37} & \meanStd{91.11}{7.78} & \meanStd{66.45}{9.08} & \meanStd{70.05}{11.32} & \meanStd{93.89}{17.32}

            \\
            \textbf{Neural}            & \meanStd{1.28}{0.49} & \meanStd{2.27}{0.98} & \meanStd{2.36}{1.43} & \meanStd{6.26}{1.87} & \meanStd{9.31}{1.00} & \meanStd{9.14}{0.77} & \meanStd{7.76}{0.81} & \meanStd{9.91}{1.29} & \meanStd{11.26}{1.08}   \\
            \midrule
            \textbf{TF}          & \meanStd{2.42}{0.52} & \meanStd{5.02}{2.24} & \meanStd{5.51}{1.83} & \meanStd{10.33}{1.85} & \meanStd{14.59}{1.01} & \meanStd{17.39}{1.13} & \meanStd{15.55}{1.54} & \meanStd{26.52}{6.95} & \meanStd{42.61}{7.57}   \\
            \bottomrule
        \end{tabular}

        \caption{Results of the over-parametrized models on representation-based \ngram LMs.}
        \label{tab:over-parametrization}        
        \end{subtable}
        \begin{subtable}{\textwidth}
        \centering
        \begin{tabular}{ccccccccccccccccc}
            \toprule
                    $\ngr$            & \multicolumn{3}{c}{$4$} & \multicolumn{3}{c}{$8$} & \multicolumn{3}{c}{$12$}                                                                                                                                                                                                                   \\
            \cmidrule{2-4} \cmidrule(lr){5-7} \cmidrule(lr){8-10} 
                            $\nsymbols$ & $64$ & $128$ & $256$ & $64$ & $128$ & $256$ & $64$ & $128$ & $256$ \\
            \midrule
            \textbf{Classic}            &  \meanStd{76.53}{24.55} & \meanStd{84.94}{42.59} & \meanStd{87.56}{9.99} & \meanStd{22.41}{8.48} & \meanStd{54.95}{4.17} & \meanStd{67.64}{5.25} & \meanStd{49.41}{8.95} & \meanStd{68.36}{5.67} & \meanStd{86.94}{6.74}  \\
            \textbf{LL}  &     \meanStd{97.80}{23.05} & \meanStd{105.58}{45.45} & \meanStd{114.81}{11.48} & \meanStd{47.93}{16.66} & \meanStd{72.46}{5.35} & \meanStd{73.03}{6.23} & \meanStd{46.98}{5.79} & \meanStd{52.33}{4.12} & \meanStd{62.85}{5.51}       \\
            \textbf{Neural}            & \meanStd{76.61}{22.81} & \meanStd{85.12}{39.41} & \meanStd{87.71}{9.32} & \meanStd{17.78}{7.90} & \meanStd{36.78}{2.82} & \meanStd{38.58}{4.74} & \meanStd{18.08}{3.03} & \meanStd{23.48}{2.94} & \meanStd{28.10}{2.16}  \\
            \midrule
            \textbf{TF}          & \meanStd{3.17}{0.62} & \meanStd{6.39}{1.39} & \meanStd{8.21}{1.84} & \meanStd{28.03}{13.64} & \meanStd{65.44}{6.39} & \meanStd{69.86}{7.50} & \meanStd{50.79}{10.58} & \meanStd{65.81}{6.71} & \meanStd{81.96}{7.43} \\
            \bottomrule
        \end{tabular}
        \caption{Results of the under-parametrized models on representation-based \ngram LMs.}
        \label{tab:under-parametrization}
        \end{subtable}
    \end{table*}

    \end{document}